\documentclass[10pt,journal,compsoc]{IEEEtran}
\ifCLASSOPTIONcompsoc
  \usepackage[nocompress]{cite}
\else
  \usepackage{cite}
\fi

\usepackage{textcomp}
\usepackage{multirow}
\usepackage{graphicx}

\usepackage[numbers,sort]{natbib}

\usepackage[normalem]{ulem}
\usepackage{subfigure}
\usepackage{url}
\usepackage{booktabs}
\usepackage{cite}
\usepackage{ifsym}
\usepackage{comment}
\usepackage{times}
\usepackage{soul}
\usepackage[hidelinks]{hyperref}
\hypersetup{
    colorlinks=true,
    linkcolor=red,
    citecolor=green,
}
\usepackage[utf8]{inputenc}
\usepackage[small]{caption}
\usepackage{amsmath}
\usepackage{amsthm}
\usepackage{algorithm}
\usepackage{algorithmic}
\usepackage[switch]{lineno}
\usepackage{tikz}
\usepackage[edges]{forest}
\usepackage{subfigure}

\renewcommand{\IEEEbibitemsep}{0pt plus 0.5pt}
\makeatletter
\IEEEtriggercmd{\reset@font\normalfont\fontsize{6.55pt}{6.55pt}\selectfont} % !!!! need to be removed later
\makeatother
\IEEEtriggeratref{1}

\begin{document}
\title{Graph Foundation Models: 

Concepts, Opportunities and Challenges}

 \author{Jiawei~Liu*,~Cheng~Yang*,~Zhiyuan~Lu,~Junze~Chen,~Yibo~Li,
 ~Mengmei~Zhang,~Ting~Bai,~Yuan~Fang,~Lichao~Sun,~Philip S.~Yu,~and~Chuan~Shi 
 \IEEEcompsocitemizethanks{
 \IEEEcompsocthanksitem Jiawei~Liu,~Cheng~Yang,~Zhiyuan~Lu,~Junze~Chen,~Yibo~Li,~Ting~Bai and Chuan~Shi are with School of Computer Science, Beijing University of Posts and Telecommunications, Beijing, China. 
%\protect\\
% % note need leading \protect in front of \\ to get a newline within \thanks as
% % \\ is fragile and will error, could use \hfil\break instead.
E-mail: \{liu\_jiawei, yangcheng, luzy, junze, yiboL, baiting, shichuan\}@bupt.edu.cn

% %\IEEEcompsocthanksitem J. Liu is with Singapore University of Technology and Design, Singapore.% <-this % stops an unwanted space
%\protect\\
% %E-mail: jun\_liu@sutd.edu.sg.

\IEEEcompsocthanksitem Mengmei Zhang is with China Telecom Bestpay, Beijing, China. 
%\protect\\
E-mail: zhangmengmei@bestpay.com.cn

\IEEEcompsocthanksitem Yuan Fang is with School of Computing and Information Systems, Singapore Management University, Singapore. 
%\protect\\
E-mail: yfang@smu.edu.sg

\IEEEcompsocthanksitem Lichao Sun is with Lehigh University, Bethlehem, Pennsylvania, USA. 
%\protect\\
E-mail: lis221@lehigh.edu

\IEEEcompsocthanksitem Philip S.~Yu is with University of Illinois Chicago, Chicago, USA. 
%\protect\\
E-mail: psyu@uic.edu

\IEEEcompsocthanksitem Jiawei Liu and Cheng~Yang contributed equally to this research.

\IEEEcompsocthanksitem Corresponding author: Chuan~Shi

% \thanks{
% ~~~~This work is supported by National Research Foundation, Singapore under its AI Singapore Programme (AISG Award No: AISG-100E-2020-065), and SUTD SRG. This work is also supported by TAILOR, a project funded by EU Horizon 2020 research and innovation programme under GA No 952215.
% }

% \thanks{
% \textcolor{blue}{~~~~This manuscript has been accepted by IEEE Transactions on Pattern Analysis and Machine Intelligence (TPAMI) - DOI: \href{https://doi.org/10.1109/TPAMI.2022.3183112}{10.1109/TPAMI.2022.3183112}. This is the up-to-date version (updated in June 2022). We plan to update this arXiv version yearly to cover the latest advances in the field of human action recognition.}
}

%\thanks{
%\textcolor{magenta}
%{
%\textcolor{magenta}{IEEE Transactions on Pattern Analysis and %Machine Intelligence (TPAMI). 
%DOI: \href{https://doi.org/10.1109/TPAMI.2022.3183112}{10.1109/TPAMI.2022.3183112}.}
%}
%}
}

% \thanks{
% ~~~~This work is supported by National Research Foundation, Singapore under its AI Singapore Programme (AISG Award No: AISG-100E-2020-065), and SUTD SRG. This work is also supported by TAILOR, a project funded by EU Horizon 2020 research and innovation programme under GA No 952215.
% }

% \thanks{
% \textcolor{blue}{~~~~This manuscript has been accepted by IEEE Transactions on Pattern Analysis and Machine Intelligence (TPAMI) - DOI: \href{https://doi.org/10.1109/TPAMI.2022.3183112}{10.1109/TPAMI.2022.3183112}. This is the up-to-date version (updated in June 2022). We plan to update this arXiv version yearly to cover the latest advances in the field of human action recognition.}
% }

\markboth{IEEE Transactions on Pattern Analysis and Machine Intelligence}% IEEE Transactions Template on Pattern Analysis and Machine Intelligence
{Graph Foundation Models: Concepts, Opportunities and Challenges}

\IEEEtitleabstractindextext{%
\begin{abstract}
Foundation models have emerged as critical components in a variety of artificial intelligence applications, and showcase significant success in natural language processing and several other domains. Meanwhile, the field of graph machine learning is witnessing a paradigm transition from shallow methods to more sophisticated deep learning approaches. The capabilities of foundation models {in generalization and adaptation} motivate graph machine learning researchers to discuss the potential of developing a new graph learning paradigm. This paradigm envisions models that are pre-trained on extensive graph data and can be adapted for various graph tasks. Despite this burgeoning interest, there is a noticeable lack of clear definitions and systematic analyses pertaining to this new domain. To this end, this article introduces the concept of Graph Foundation Models (GFMs), and offers an exhaustive explanation of their key characteristics and underlying technologies. We proceed to classify the existing work related to GFMs into three distinct categories, based on their dependence on graph neural networks and large language models. In addition to providing a thorough review of the current state of GFMs, this article also outlooks potential avenues for future research in this rapidly evolving domain.

%In contrast to the bottlenecks faced by traditional graph deep learning models in terms of expressive power and generalization, graph foundation models are expected to emerge with stronger capabilities and achieve universality across diverse tasks.
\end{abstract}

}

\maketitle

\IEEEdisplaynontitleabstractindextext

% \fi
\IEEEpeerreviewmaketitle

\renewcommand{\baselinestretch}{0.9} % !!!! need to be removed later !!!!!!!!!!

\section{Introduction}
\IEEEPARstart{W}{ith} the rise in computational power and breakthroughs in deep learning techniques, the artificial intelligence (AI) community has introduced the notion of ``foundation models'': \textit{A foundation model is any model that is trained on broad data and can be adapted to a wide range of downstream tasks~\cite{bommasani2021opportunities}.} Foundation models enjoy unique attributes like emergence and homogenization, empowering them to serve as the primary building blocks for a myriad of downstream AI applications~\cite{bommasani2021opportunities}. Emergence suggests that as a foundation model scales up, it may spontaneously manifest novel capabilities~\cite{wei2022emergent}. Meanwhile, homogenization alludes to the model's versatility, enabling its deployment across diverse applications~\cite{bommasani2021opportunities}. Thanks to the development of large language models (LLMs), the concept of foundation models first became a reality in natural language processing (NLP). Since then, foundation models have demonstrated impressive versatility, processing not just text but also image data, video data, audio data and multi-modal inputs. This versatility empowers them to excel in tasks ranging from computer vision~\cite{wang2023visionllm} and audio signal processing~\cite{zhang2023video} to recommender systems~\cite{fan2023recommender}.

Much like the evolution witnessed in NLP, graph machine learning is also undergoing a paradigm transition. {In its early stages, graph tasks predominantly employed shallow methods, such as random walk~\cite{DeepWalk,grover2016node2vec} and matrix factorization~\cite{yan2006graph,trouillon2016complex,nickel2011three,yang2015network,wang2022duality}.} These methods, however, were typically limited to transductive learning~\cite{zhou2020graph}. The more recent shift towards deep learning methods has catalyzed the rise of graph neural networks (GNNs). GNNs have revolutionized the landscape by introducing the message-passing mechanism, where nodes iteratively aggregate information from their neighbors. By harnessing GNNs in fully supervised, semi-supervised, or unsupervised settings, researchers have pioneered a variety of customized graph models. These advancements have yielded substantial improvements in tasks like node classification~\cite{kipf2016semi}, link prediction~\cite{zhang2018link}, graph classification~\cite{xu2018powerful}, and graph clustering~\cite{wang2019attributed}. However, certain challenges of GNN models still persist. For example, GNNs are restricted with issues related to expressive power~\cite{alon2020bottleneck} and generalizability~\cite{yang2023individual}, especially given the ever-expanding datasets and the widening spectrum of tasks.

% Similar to NLP, graph machine learning is also undergoing a shift in model paradigms. In the early stages, graph-based tasks were addressed using shallow methods, such as random walk and matrix factorization, which can only support transductive learning on unattributed graphs~\cite{waikhom2021graph}. In recent years, the pivotal shift towards deep learning architectures has led to the emergence of graph neural networks (GNNs), which introduced the concept of message passing, allowing nodes to gather information from their adjacent nodes iteratively. Training GNNs in a fully supervised, semi-supervised, or unsupervised manner, researchers have developed a wide range of customized graph models, which have led to significant advancements in tasks like node classification~\cite{kipf2016semi}, link prediction~\cite{zhang2018link}, graph classification~\cite{xu2018powerful} and graph clustering~\cite{wang2019attributed}. However, these models still face bottlenecks in terms of expressive power~\cite{alon2020bottleneck} and generalization~\cite{yang2023individual}, making it challenging for them to cope with the ever-growing data and increasingly diverse tasks.
% 介绍训练范式：unsupervised/semi-supervised/full-supervised

% 1. 我们期待变高，通过预训练+大量数据，模型能力提升。 2. 期待适用于各种任务。
\begin{figure}[!ht]
    \centering
    \includegraphics[width=0.99\linewidth]{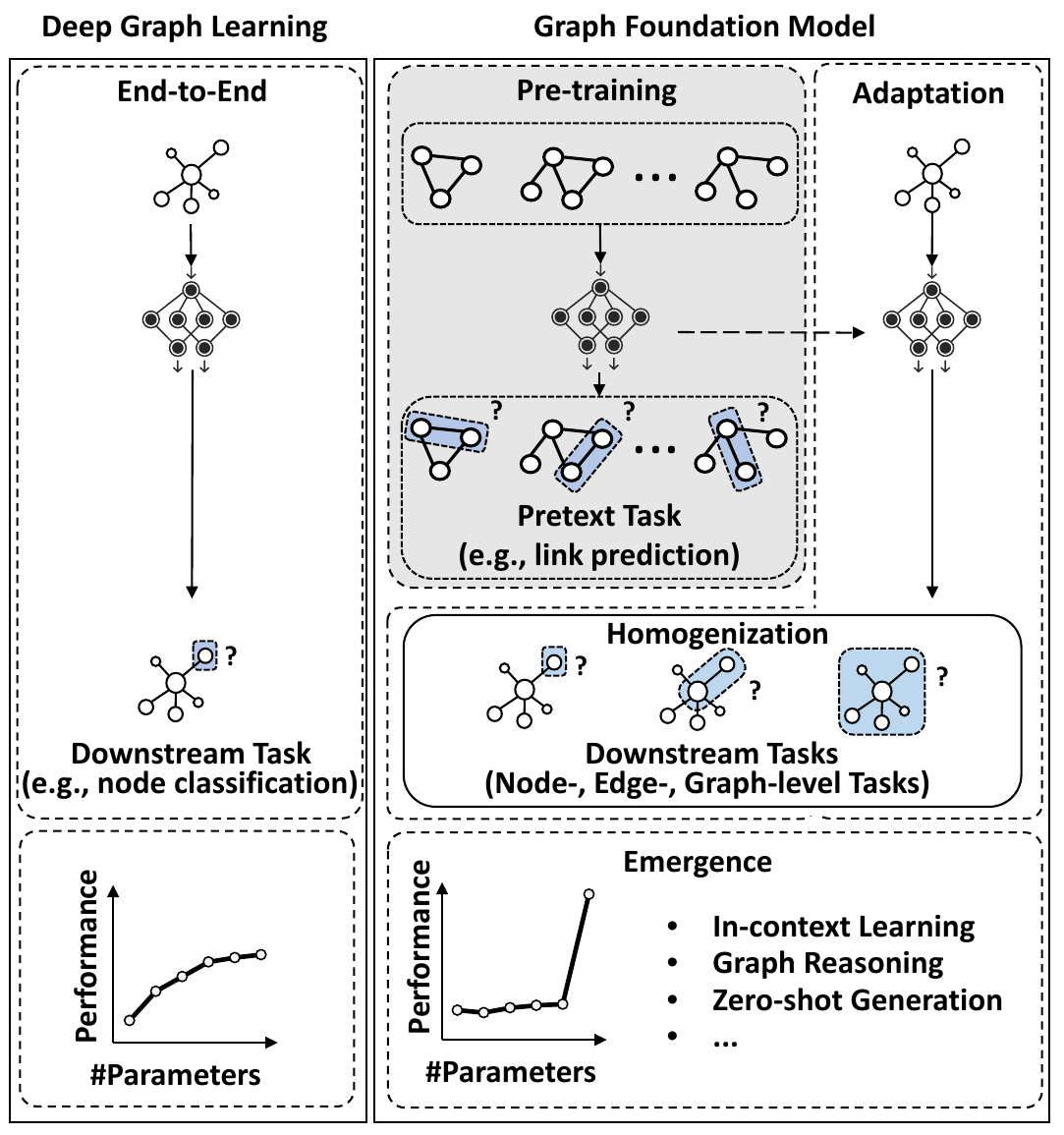}
    \caption{The distinction between deep graph learning and graph foundation models. Deep graph learning tackles specific tasks on specific datasets through end-to-end training. %In contrast, foundation models are initially pre-trained on extensive datasets using pretext tasks and subsequently adapted to new datasets and diverse tasks, exhibiting homogenization. As the model's parameter count increases, foundation models can also emerge with new capabilities, such as in-context learning, graph reasoning, and zero-shot generation.
    In contrast, graph foundation models (GFMs) are pre-trained on broad graph data and can be adapted to a wide range of downstream graph tasks, expected to demonstrate emergence and homogenization capabilities.
    }
    \label{fig:intro}
\end{figure}

The remarkable success of foundation models in varied domains is increasingly garnering the interest of graph machine learning researchers. This naturally evokes the question: {Could graph foundation models represent the next frontier in graph machine learning?} Such models, if realized, would boast enhanced expressive power, improved transferability, and applicability to more intricate graph data and tasks. As illustrated in Figure~\ref{fig:intro}, a graph foundation model (GFM) is envisioned as a model pre-trained on extensive graph data, primed for adaptation across diverse downstream graph tasks. Drawing parallels with traditional foundation models, a GFM is also anticipated to embody two principal characteristics: emergence and homogenization. Specifically, emergence refers to novel capabilities shown exclusively in {large-scale} graph models, while homogenization denotes the model's adaptability across {different} types of graph tasks. Existing deep graph learning methods struggle to encompass these features: their inherent architectures and learning paradigms focus on specific tasks, which restrict the utilization of extensive unlabeled data, subsequently limiting their expressive and generalization abilities.

%  
% The remarkable success of foundation models in other domains is capturing the attention of graph machine learning researchers. It inevitably leads us to ponder whether graph foundation models will become the next-generation paradigm for graph machine learning, possessing stronger expressive power and transferability, and applicable to more challenging graph data and tasks. As illustrated in Figure~\ref{fig:intro}, 
% %\textit{a graph foundation model (GFM) is a model that is trained on broad graph data and can be adapted to a wide range of downstream graph tasks.} 
% \textit{a graph foundation model (GFM) is a model that is expected to benefit from the pre-training of broad graph data, and can be adapted to a wide range of downstream graph tasks.}
% %Similar to foundation model, graph foundation model is not new from a technical perspective: graph neural networks and self-supervised learning are its predecessors. 
% %A graph foundation model (GFM) is expected to be trained on broad graph data, and can adapt to a wide range of downstream graph tasks.
% Similar to foundation model, GFM is also expected to have two key characteristics: emergence and homogenization, namely the new abilities only present in large-scale graph models and the ability to be widely applicable across various graph tasks. Existing graph models lack these capabilities because their backbone architectures and learning paradigms are challenging to leverage large-scale unlabeled data, resulting in limited expressive power and generalization ability of the models.

Inspired by the success of LLMs as foundation models in NLP, researchers have explored the possibilities of graph foundation models towards the emergence and homogenization capabilities.
%Inspired by the success of LLMs as foundation models for NLP, some researchers have attempted to draw from related techniques to explore GFMs, aiming to achieve emergence and homogenization capabilities. 
These explorations primarily focus on the design of backbone architectures for GFMs, and different learning paradigms including pre-training and adaptation, as they are the key strategies of LLMs to acheive the aforementioned capabilities.
First and foremost, the emergent abilities of foundation models typically exist only in backbones with a large number of parameters, whereas the parameter count of GNNs is significantly smaller than that of the language backbones. This implies that the backbone of GFMs may need to be redesigned to achieve more substantial knowledge storage towards emergence. As graph data is typically associated with rich text information, an alternative approach is to use LLMs as GFMs. Nonetheless, it remains uncertain whether LLMs can effectively handle graph data and associated tasks, and it is crucial to determine how to model graph structures in LLMs. Additionally, the homogenization of foundation models necessitates the handling of diverse tasks in a uniform manner. Devising effective pre-training tasks (also called pretext tasks) and downstream task adaptation methods are challenging for graph data, due to the complexity in interconnected nodes and various forms of attributes, as well as the diversity in tasks across node-, edge- and graph-levels. Therefore, there is also a need to design suitable pre-training tasks and adaptation mechanisms.

%  the emergence capability relies on prompts to stimulate it. In contrast to words that possess rich semantic information, the basic elements in graphs are abstract nodes and edges. Additionally, designing prompts for graphs becomes challenging due to the abstract nature of graph structures. 

% 凸显两个Challenges：1. transferability 2. substantial knowledge storage capabilities

While there is no definitive solution for designing and implementing GFMs, this paper surveys some related researches and categorizes them into three distinct approaches based on their reliance on GNNs and LLMs: (1) 
\textbf{GNN-based Models}: They aim to enhance existing graph learning paradigms through innovations in the backbone, pre-training, and adaptation aspects; (2) \textbf{LLM-based Models}: 
They explore the feasibility of using an LLM as a GFM by converting graphs into text or tokens; (3) \textbf{GNN+LLM-based Models}: They explore various forms of synergy between GNNs and LLMs to empower them with enhanced capabilities. 
%The basis for this taxonomy is that each category of methods possesses distinct backbones and corresponding training strategies, which can be viewed as different solutions to the two challenges.
%Finally, we explore several potential research directions for graph foundation models in this article. 

% \begin{figure}[h!]
%     \centering
%     \includegraphics[width=0.95\linewidth]{images/backbone-Taxonomy.pdf}
%     \caption{Taxonomy of GFMs, based on their backbone architectures. Standalone GM and standalone LLM only utilize a single graph model (GM) or a large language model (LLM) as GFM, and synergized LLM+GM combines GM and LLM in various ways as GFM.}
%     \label{fig:backbone}
% \end{figure}

To the best of our knowledge, this is the first survey towards graph foundation models. Existing surveys of foundation models typically explore modalities such as language and vision~\cite{bommasani2021opportunities,zhou2023comprehensive}, rather than graphs. Additionally, there are two surveys~\cite{pan2023unifying,pan2023large} dedicated to knowledge graphs and large language models, but knowledge graphs, due to their distinct nature in construction and application, fall outside the scope of this article. We have also noticed a very recent article that mentions the concept of large graph models~\cite{zhang2023large}, but it emphasizes opinion statements and lacks a systematic taxonomy. Therefore, the contributions of this article can be summarized as follows:

$\bullet$ This article defines the concept of graph foundation models for the first time, and examines the core issues and characteristics of their capabilities.

$\bullet$ This article introduces a novel taxonomy and discusses the strengths and limitations of each approach towards graph foundation models.

$\bullet$ This article provides promising future directions towards graph foundation models.

The subsequent sections are organized as follows. In Section~\ref{sec:Background}, we introduce the background related to GFMs. Section~\ref{sec:Characteristics} defines GFMs and highlights their similarities and differences with language foundation models. Sections~\ref{sec:GM} - \ref{sec:LLM+GM} delve into the relevant works that consider GNN-based models, LLM-based models and GNN+LLM-based models as GFMs, separately. Section~\ref{sec:future} engages in a discussion on the future directions of GFMs. In Section~\ref{sec:conclusions}, we summarize the key points of this paper.

\section{Background}
\label{sec:Background}
% Before introducing the concept of graph foundation models, in this section, we first review some background knowledge, namely, deep graph learning and language foundation models. Specifically, we introduce them from three aspects: data, backbone architectures, and learning paradigms.
Before introducing GFMs, we review background knowledge on deep graph learning and language foundation models.%, focusing on data, backbone architectures, and learning paradigms.

\subsection{Deep Graph Learning}
%Many real-world systems are naturally represented as graphs. Deep graph learning is crucial for modeling complex relationships across various fields. 
This section provides a concise overview from three key aspects: graph data, backbone architectures, and learning paradigms.

\subsubsection{Graph Data} 

{Graphs capture intricate relationships among entities and possess several key characteristics that make them challenging for machine learning tasks. The primary challenge stems from their \textbf{(1) Non-Euclidean Nature}: Graph data is inherently non-Euclidean~\cite{wu2020comprehensive}, lacking the rigid geometric structure of traditional data formats such as 1D text, 2D images or tabular data. This means that graph data cannot be adequately described within a simple flat space because its intrinsic structure does not conform to the principles of Euclidean geometry~\cite{zafeiriou2022guest}. Unlike Euclidean data, which often comes in predefined shapes (e.g., images of a specific resolution), non-Euclidean data can vary greatly in size and shape. This variability complicates the design of algorithms that often navigate complex topologies, significantly increasing computational cost compared to operations on simpler Euclidean spaces.}

Beyond this fundamental structural complexity, two additional challenges are posed by the nature of graph data. \textbf{(2) Various Domains}: Graph data appears in domains such as social networks~\cite{freeman2004development}, biology~\cite{muzio2021biological}, and transportation~\cite{jiang2022graph}. It is also used in tasks like 3D human skeleton recognition~\cite{li2021symbiotic}, semantic segmentation~\cite{zhang2021affinity}, and video classification~\cite{gao2020learning}. Domain-specific variability with different node types and edge semantics makes creating a universal model challenging~\cite{fan2023generalizing}. \textbf{(3) Various Types}: graph data includes homogeneous, heterogeneous~\cite{shi2016survey}, hyper-~\cite{feng2019hypergraph}, and dynamic ones~\cite{lu2024motion}. Such diversity also brings challenges to deep graph learning.

\subsubsection{Backbone Architectures}
GNNs are the current mainstream backbone architecture for deep graph learning. Most GNNs follow the message-passing framework \cite{gilmer2017neural}, enabling nodes to exchange information with neighbors. For example, GCN~\cite{kipf2016semi} introduces graph convolution layers, GraphSAGE~\cite{hamilton2017inductive} generates node embeddings using inductive learning, and GAT~\cite{velivckovic2017graph} adds an attention mechanism to weigh neighbor importance, enhancing expressive power. These contributions make GNNs versatile tools for deep graph learning.

However, deepening GNNs is challenging. Increasing layers leads to over-smoothing, where node representations become similar~\cite{li2018deeper}, and over-squashing, where information is overly compressed~\cite{alon2020bottleneck}. Efforts like DropEdge~\cite{rong2019dropedge}, which randomly removes edges, improve GCN performance and scalability. Graph transformers~\cite{ying2021transformers,chen2022structure,kreuzer2021rethinking}, with their fully connected attention and long-range relationship modeling, help alleviate over-smoothing and over-squashing~\cite{rampavsek2022recipe}.

\subsubsection{Learning Paradigms}
The learning paradigms for deep graph learning encompass three primary categories:%s supervised, semi-supervised, and unsupervised learning. We will briefly introduce these paradigms.

\textbf{Supervised learning.}
In supervised learning, algorithms use a training dataset with input data and output labels. This is used in tasks like graph classification~\cite{lee2018graph} and graph regression~\cite{jiang2019censnet}. For instance, in molecular property prediction~\cite{wieder2020compact}, GNNs predict chemical properties using labeled data, aiding drug development and materials research.

\textbf{Semi-supervised learning.}
Semi-supervised learning, as discussed in \cite{song2022graph}, is a primary focus in deep graph learning. It uses both labeled and unlabeled data to improve model performance, with node classification \cite{kipf2016semi} being a key application. The message-passing mechanism \cite{gilmer2017neural} allows GNNs to exchange information among nodes, incorporating both data types for predictions. GNNs can also combine with methods like label propagation for better performance~\cite{yang2021extract}.

\textbf{Unsupervised learning.}
Unsupervised learning discovers patterns and structures without manual labels. Graph clustering identifies structures based on node relationships, while link prediction estimates missing connections. A subclass, self-supervised learning, generates labels from the data itself~\cite{liu2022graph}, allowing GNNs to be trained end-to-end for tasks like graph clustering~\cite{wang2019attributed} and link prediction~\cite{zhang2018link}.
%~\cite{rahman2021comprehensive}

\subsection{Language Foundation Models}
AI is currently undergoing a transformative shift marked by the emergence of some specific language models (such as GPT-3) that are trained on extensive and diverse datasets using self-supervised learning. These models, known as {foundation models}, are able to produce a wide array of outputs, enabling them to tackle a broad spectrum of downstream tasks. In contrast to the deep graph learning pipeline, the foundation model's approach embraces a {pre-training and adaptation} framework, enabling the model to achieve several significant advancements, including the emergence \cite{wei2022emergent} and homogenization \cite{bommasani2021opportunities}.
% 加引用
%Foundation models have primarily solidified their presence in NLP \cite{bommasani2021opportunities}, so, for now, our narrative will be centered around this domain.
Foundation models have primarily established themselves in the field of NLP~\cite{bommasani2021opportunities}, so our discussion will focus on language foundation models in this section.

% add by zmm: 可以参考这里的逻辑叙述，引出pretrain和adaptation：
% 基础模型概念之所以重要，正是因为它实现了人工智能范式的转变。这种范式转变突出了模型本身的多功能性和适应性，不再局限于单一任务或领域。这也意味着，为了有效地适应各种下游任务，不一定需要从头开始设计一个新模型或算法。相反，我们可以从一个预先训练好的基础模型开始，然后对其进行微调，以适应特定的需求或应用。这不仅大大提高了效率，也推动了AI技术的广泛应用。

\subsubsection{Language Data}
% 第一段定义，第二段数据特点
Language data refers to text or spoken content in a human language, encompassing the grammar rules of the natural language and the associated semantics of the words. 
% It can include written documents, transcribed audio recordings, and any other form of language-based communication. Language data is essential for many NLP tasks, such as machine translation, sentiment analysis and text summarization. Researchers and developers use language data to train and evaluate language models and other NLP algorithms.
The quality and quantity of language data play a crucial role in the performance of NLP systems, impacting their accuracy, robustness, and overall effectiveness in various language tasks. 
% In contrast to computer vision and other domains, the size of annotated language data is rather small, consisting of only a few thousand sentences \cite{paass2023foundation}. This limitation is primarily due to the high cost of manual annotation. Nevertheless, there is a vast amount of unlabeled language data available from sources such as the internet, newspapers, and books, creating opportunities for utilizing unlabeled data in model pre-training. 
In contrast to graph data, language data as Euclidean data is easier to model, and its rich semantic information significantly enhances the knowledge transferability of language models.

%The availability of this data can be attributed to the fact that natural language is one of the oldest and most universal means of human data recording.

\subsubsection{Backbone Architectures}

% PLM
An early breakthrough in foundation models is pre-trained language models (PLMs), designed to capture context-aware word representations, which proved remarkably effective as versatile semantic features. 
% For instance, BERT \cite{devlin2018bert}, grounded in the parallelizable Transformer architecture \cite{vaswani2017attention} with self-attention mechanisms, is conceived through the pre-training of bidirectional language models with specifically designed pretext tasks on vast unlabeled data. This landmark study significantly elevates the performance benchmarks for NLP tasks and serves as a catalyst for a plethora of subsequent research, establishing the prevailing pre-training and fine-tuning learning paradigm.
%BERT [14], based on the highly parallelizable Transformer architecture [121] with self-attention mechanisms, is developed through the pre-training of bidirectional language models using specifically designed pretext tasks on a large amount of unlabeled data.
% LLM, 介绍一下token 
Furthermore, researchers have observed that increasing the scale of PLMs, whether by augmenting model size or training data, frequently results in increased model capacity for downstream tasks. These larger PLMs, collectively referred to as LLMs, exhibit emergent abilities \cite{wei2022emergent} compared to their smaller counterparts (e.g., the 1.5B-parameter GPT-2 and 175B-parameter GPT-3). 
% After training on massive text datasets, they manifest remarkable capabilities, often referred to as emergent abilities \cite{wei2022emergent}, such as in-context learning \cite{bommasani2021opportunities}. 
% Transformer 介绍
LLMs primarily utilize the Transformer architecture, because highly parallelizable Transformer-based architectures accelerate the pre-training stage and enable the utilization of massive datasets. In Transformer models, tokens serve as the input and represent units at the word level in natural language texts. 
% Typically, LLMs containing hundreds of billions (or more) of parameters \cite{zhao2023survey}, exemplified by models such as GPT-3 \cite{brown2020language}, PaLM \cite{PaLM}, Galactica \cite{Galactica}, and LLaMA \cite{LLaMA}.

% Presently, LLMs wield a profound influence on the AI community and are increasingly recognized as a highly promising choice to serve as the backbone for foundation models in other modalities, such as computer vision~~\cite{wang2023visionllm} and audio signal processing~\cite{zhang2023video}. 
%Currently, LLMs have a significant impact on the AI community and are increasingly acknowledged as a highly promising choice for serving as the backbone of foundation models in other domains, including computer vision \cite{wang2023visionllm} and audio signal processing \cite{zhang2023video}.

\subsubsection{Learning Paradigms}
\label{sec:background-llm-pretraining}
%什么是预训练，利用unlabeled data
As the number of model parameters has rapidly increased, the demand for significantly larger datasets has grown to effectively train these parameters and avoid overfitting. Given the extremely expensive costs associated with building large-scale labeled datasets, the importance of utilizing extensive unlabeled text data has been underscored. Leveraging these unlabeled datasets involves a two-step approach: first, achieving universal representations through self-supervised learning, and subsequently employing these representations for various tasks \cite{qiu2020pre}. Based on different adaptation approaches, learning paradigms can be categorized into two types: pre-train and fine-tune and pre-train, prompt, and predict~\cite{liu2023pre}.

\textbf{Pre-train and Fine-tune}.
% 预训练的优势。优点(：下游任务不绑定；自监督训练->可以利用无标签的数据)
In this paradigm, a model is initially pre-trained as a language model (LM), where it predicts the probability of observed textual data.
Following the pre-training phase, 
% foundation models acquire general-purpose capabilities suitable for a broad spectrum of tasks. Nevertheless, pre-trained models still lack downstream task-specific information, and using them directly may not yield optimal results. Therefore, 
we need to tune the model for specific tasks, which is known as fine-tuning. Building upon the success of models like ULMFit \cite{howard2018universal} and BERT \cite{devlin2018bert}, fine-tuning has emerged as the predominant method for adapting pre-trained models. In this framework, the primary emphasis lies in objective engineering, encompassing the design of training objectives for both pre-training and fine-tuning phases. 
% For instance, Pegasus~\cite{zhang2020pegasus} shows that incorporating a loss function for predicting important sentences within a document results in an improved pre-trained model for text summarization.
% Within this paradigm, the focus turned mainly to objective engineering, designing the training objectives used at both the pre-training and fine-tuning stages. For example, Pegasus~\cite{zhang2020pegasus} shows that introducing a loss function of predicting salient sentences from a document will lead to a better pre-trained model for text summarization.
% fine-tune的优点：可以弥补预训练和下游任务的gap(transfer)
The advantage of fine-tuning is that it can transfer knowledge between source and target tasks (or domains) and benefit the model's performance. For the small size of fine-tuning dataset compared to pre-training dataset, this process can enable adaptation effectively without losing much pre-trained knowledge.
% Despite the relatively small size of the fine-tuning dataset compared to the extensive pre-training data, the model's high capacity, with millions of parameters, enables effective adaptation without compromising the structural language knowledge it has acquired. This concept has been proven applicable to a wide array of NLP tasks, resulting in unprecedented improvements in semantic understanding. 
%It's worth noting, however, that the fine-tuning process can be intricate, and even with identical hyperparameter settings, different random seeds may produce significantly divergent outcomes.

\textbf{Pre-train, Prompt and Predict}.
% In this paradigm, instead of adapting pre-trained LMs to downstream tasks via objective engineering, downstream tasks are reformulated to look more like those solved during the original LM training with the help of a textual prompt. In this way, by selecting the appropriate prompts we can manipulate the model behavior so that the pre-trained LM itself can be used to predict the desired output, sometimes even without any additional task-specific training. The advantage of this method is that, given a suite of appropriate prompts, a single LM trained in an entirely unsupervised fashion can be used to solve a great number of tasks. \cite{liu2023pre}
In this paradigm, rather than adjusting PLMs to suit specific downstream tasks, the approach involves reshaping the downstream tasks to align more closely with those tackled during the original LM training, accomplished by providing textual prompts. 
% By selecting suitable prompts, we can steer the LM's behavior so that it can predict the desired output, sometimes without any additional task-specific training. This method offers the advantage of enabling a single, entirely unsupervised LM, when equipped with a set of fitting prompts, to handle a wide array of tasks \cite{liu2023pre}.
% prompt的优势，除了fine-tune之外，prompt是另一类方法。或者，从fine-tune的缺点引到prompt：引着下游任务更贴近pretrain任务。
% Another way to reduce the disparity between pre-training and downstream task is reformulating the downstream tasks by designing appropriate prompts, so called prompting \cite{liu2023pre}. In this process, downstream tasks are adapted to resemble the tasks LLMs encountered during the original training phase, enhancing their performance on downstream tasks. 
% 从prompt engineering的角度，可分为maual 和automated
From the aspect of prompt engineering, the approaches to create a proper prompts can be classified to manual methods and automated methods. Manual methods involve creating intuitive templates based on human insight, which is the most straightforward approach to crafting prompts. 
% For instance, the influential LAMA dataset \cite{petroni2019language} offers manually devised cloze templates to assess the knowledge of language models. 
Manual methods face challenges in terms of high cost and precision. To address these issues, some approaches have started to experiment with automated prompt generation. 
When LLMs have billions of parameters, it is more effective if we can just adapt the downstream tasks without adjusting model parameters. This makes prompt-tuning a promising approach for adapting LLMs.
\section{Graph foundation models}
\label{sec:Characteristics}
\begin{table*}[ht]
\centering
\caption{\footnotesize{The relationship between language foundation model and graph foundation model.}}
\resizebox{0.95\linewidth}{!}{
\begin{tabular}{@{}llll@{}}
\toprule
                                                                                 &                    & Language Foundation Model              & Graph Foundation Model                                                                                                                           \\ \midrule 
\multirow{2}{*}{Similarities}                                                    & Goal               & \multicolumn{2}{l}{\begin{tabular}[l]{@{}c@{}}\multicolumn{1}{c}{Enhancing the model's expressive power and its generalization across various tasks}\end{tabular}}                                         \\ \cmidrule(l){2-4}
                                                                                 & Paradigm           & \multicolumn{2}{c}{Pre-training and Adaptation}                                                                                                                                                \\ \midrule
\multirow{2}{*}{\begin{tabular}[c]{@{}c@{}}Intrinsic\\ differences\end{tabular}} & Data               & Euclidean data (text)                  & \begin{tabular}[l]{@{}l@{}}Non-Euclidean data (graphs) or a mixture of Euclidean \\ (e.g., graph attributes) and non-Euclidean data\end{tabular} \\ \cmidrule(l){2-4}
                                                                                 & Task               & Many tasks, similar formats              & Limited number of tasks, diverse formats                                                                                                                         \\ \midrule
\multirow{4}{*}{\begin{tabular}[c]{@{}c@{}}Extrinsic\\ differences\end{tabular}} & Backbone Architectures    & Mostly based on Transformer            & No unified architecture                                                                                                                        \\ \cmidrule(l){2-4}
                                                                                 & Homogenization     & Easy to homogenize      & Difficult to homogenize                                                                                                                          \\ \cmidrule(l){2-4}
                                                                                 & Domain Generalization & Strong generalization capability       & Weak generalization across datasets                                                                                                              \\ \cmidrule(l){2-4}
                                                                                 & Emergence          & Has demonstrated emergent abilities & No/unclear emergent
                                                                                 abilities as of the time of writing                                                                                                                \\ \bottomrule
\end{tabular}}

\label{tab:relationship}
\end{table*}
In this section, we will first formally define the concepts of graph foundation models, including the definition, key characteristics and key technologies. Then, we will discuss the similarities and differences between graph and language foundation models.%Then, we will discuss the impact from graph data and graph tasks on graph foundation models. Finally, we will discuss the similarities and differences between graph foundation models and language foundation models.

\subsection{Definition and Key Characteristics}
We define a graph foundation model as follows:

\textit{\textbf{Definition}} A graph foundation model (GFM) is a model that is expected to benefit from the pre-training of broad graph data, and can be adapted to a wide range of downstream graph tasks.

% \textsc{Definition}:

Compared to deep graph learning that employs end-to-end training, GFMs use {pre-training} to obtain the knowledge from a substantial amount of unlabeled graph data, and then use {adaptation} techniques to tailor to various downstream tasks. Some studies~\cite{liu2023graphprompt,allinone} have already demonstrated that the paradigm of pre-training and adaptation outperform deep graph learning in certain scenarios, e.g., few-shot learning~\cite{liu2023graphprompt}, showcasing their superior expressive power and generalization ability. Unlike deep graph learning that aims to achieve better performance on a single task, a GFM is expected to have two key characteristics: emergence and homogenization.

\textbf{Emergence}. Emergence means that the graph foundation model will exhibit some new abilities when having a large parameters or trained on more data, which are also referred to as emergent abilities. Inspired by the various emergent abilities~\cite{dong2022survey,wei2022chain} possessed by foundation models, we expect GFMs to have similar abilities, including in-context learning, graph reasoning, and zero-shot graph generation, etc. In-context learning allows predictions for various downstream tasks with few examples~\cite{PRODIGY}. Graph reasoning decomposes a complex problem into multiple sub-problems based on the graph structure and addresses them step by step, such as solving graph algorithm problems~\cite{NLGraph}. Zero-shot graph generation requires the model to generate graphs based on the desired conditions without the need for any examples~\cite{su2022molecular}. Note that although language foundation models have demonstrated various emergent abilities, only a few works~\cite{PRODIGY,NLGraph,su2022molecular} have explored emergent abilities of GFMs so far.%li2023blip

\textbf{Homogenization}. Homogenization means that the graph foundation model can be applied to different formats of tasks, such as node classification, link prediction and graph classification. Note that due to the distinct characteristics of tasks on graphs compared to NLP tasks, achieving homogenization is not straightforward. The fundamental question in achieving homogenization is to decide which form to unify different types of graph tasks. Existing works have attempted homogenization through link prediction~\cite{liu2023graphprompt} or graph-level tasks~\cite{allinone}, but there is no consensus on which approach is superior.
% refers to the phenomenon where as the number of parameters in the backbone architecture of the graph foundation model increases, the model exhibits new capabilities, such as in-context learning, reasoning, and zero-shot generation.

\subsection{Key Technologies}
Graph foundation models primarily comprise two key techniques: pre-training and adaptation. This section will provide a brief overview of these two techniques.

% Graph: Contrastive, Predictive
% Language: LM, MLM, etc.
\textbf{Pre-training}. Pre-training is a pivotal concept in the development of graph foundation models, akin to its role in language models. It involves pre-training a neural network on a large graph dataset in a self-supervised manner. During pre-training, the model learns to capture the structural information, relationships, and patterns within the graph. There are several pre-training strategies for graph foundation models. Contrastive self-supervised learning~\cite{GRACE,sun2021mocl} leverages the idea of learning representations by contrasting positive samples (e.g., similar node pairs) against negative samples (e.g., dissimilar node pairs). Generative self-supervised learning~\cite{GraphMAE,GraphMAE2} encourages the model to reconstruct the structure or predict the features of original graph data. If using LLM as a part of the graph foundation model, we can also employ the pre-training methods introduced in Section~\ref{sec:background-llm-pretraining}. These diverse pre-training approaches enable graph foundation models to learn meaningful representations from raw graph data, enhancing their generalization and adaptability across various graph tasks. 

% Self-supervised Learning: Contrastive, Generative
%The pre-training of graph foundation models is a crucial phase in their development, analogous to the pre-training of large language models in natural language processing. During pre-training, these models are exposed to vast amounts of graph-structured data to learn rich representations and capture meaningful patterns. The process involves training the model on a massive graph dataset, where nodes represent entities, and edges signify relationships or interactions between them. Through self-supervised learning tasks, such as node attribute prediction, link prediction, or graph completion, the model learns to encode the structural information, dependencies, and context within the graph. This pre-training phase enables the model to acquire a foundational understanding of graph data, which can be fine-tuned for specific downstream tasks. The resulting pre-trained graph foundation model serves as a powerful starting point for a wide array of graph-related applications, offering the potential for improved performance and generalization.
% No Tuning:
% Fine tuning: internal
% Prompt tuning: external

% 预训练不是新概念（Why Does Unsupervised Pre-training Help Deep Learning? 2010）

% Fine-tuning: Vanilla, Parameter-efficient
% Prompt-tuning: Discrete, Continuous
\textbf{Adaptation}. The adaptation of graph foundation models involves tailoring these models to specific downstream tasks or domains to enhance their performance. This process includes several techniques, i.e., vanilla fine-tuning, parameter-efficient fine-tuning and prompt-tuning. Vanilla fine-tuning (Vanilla FT) entails training the entire pre-trained model on task-specific data, allowing for the highest level of customization but often requiring substantial data and computational resources. Parameter-efficient fine-tuning (Parameter-efficient FT)~\cite{adaptergnn,G-adapter}, on the other hand, adjusts only a subset of the model's parameters, striking a balance between task-specific adaptation and resource efficiency. Prompt-tuning~\cite{sun2022gppt,allinone} is a versatile approach that relies on external prompts to guide the model's behavior, making it more adaptable and effective. These adaptation techniques enable graph foundation models to excel in a wide range of applications by leveraging their pre-trained knowledge while tailoring their capabilities to specific tasks or domains, making them valuable for diverse downstream applications. Note that although LLMs have developed various types of prompt-tuning methods~\cite{liu2023pre} and some other efficient tuning methods, such as Prefix Tuning~\cite{li2021prefix}, there are relatively few prompt tuning methods for graph foundation models.

\subsection{Comparison between GFMs and LLMs}
\label{sec:comparison-gfm-llm}

Through conceptual comparison, we can observe similarities in the goals and learning paradigms between graph foundation models (GFMs) and language foundation models (commonly referred to as large language models, LLMs). However, the uniqueness of graph data and graph tasks creates fundamental differences between them, which we refer to as their intrinsic differences. Furthermore, due to the relatively limited research on GFMs at present, many issues that have been extensively explored in LLMs remain unresolved, which we refer to as their extrinsic differences. We summarize the similarities and differences between GFMs and LLMs in Table~\ref{tab:relationship}, and will delve into them in detail in this section.
%Language Foundation Models and graph foundation models are both powerful paradigms in the field of artificial intelligence and natural language processing, aimed at enhancing a model's expressive power and generalization across various tasks.
\subsubsection{Similarities}
As shown in Table~\ref{tab:relationship}, both language foundation models and graph foundation models share the common goal of enhancing a model's expressive power and improving its ability to generalize across a wide range of tasks. They aim to create versatile, pre-trained models that can be adapted for specific applications. In addition, both follow the pre-training and adaptation paradigm. They begin by pre-training a model on a large, diverse dataset and then adapt it to task-specific data.

%\subsection{Differences from Language Foundation Model}
%While both paradigms follow the Pretraining-Adaptation approach and aim to enhance model generalization, they target different data types and task domains, which are the intrinsic differences of them. Furthermore, due to technical limitations, graph foundation models also exhibit some extrinsic differences from Language Foundation Model.

\subsubsection{Intrinsic Differences}

%language foundation models are specialized for processing text data and are versatile across various text-related tasks. In contrast, graph foundation models are tailored for working with non-Euclidean data in the form of graphs and are well-suited for tasks related to structured data and network analysis. 
The intrinsic differences between GFM and LLM primarily manifest in two aspects: data and tasks. As for input data, language foundation models are primarily designed for processing Euclidean data, i.e., text. They are trained on vast text corpora, which are inherently sequential and follow a linear order of words or tokens. On the other hand, GFMs are designed to handle non-Euclidean data (represented as graph structures) or a mixture of Euclidean data (like graph attributes) and non-Euclidean data. Compared to text data, graph data can capture complex data relationships and is typically more sparse. Moreover, 
%as mentioned in Section~\ref{sec:impact-graph-data}, 
different graphs may exhibit significant differences in type or structure/geometry, all of which pose challenges in the design of GFMs. Furthermore, language data, even when sourced from texts in different domains, still share a common vocabulary. On the other hand, different graph data may lack this common foundation. For instance, nodes represent atoms in a molecular graph, while nodes represent users in a social network, which are entirely different. Furthermore, graphs from different domains can have different structures. Some graphs have a more hierarchical structure, while others may have a more cyclical structure~\cite{sun2022self}. Moreover, for a single graph, the different regions can exhibit different structures. For example, In recommender systems, the user-user subgraph and item-item subgraph generally exhibit very distinct structures~\cite{ye2023sincere}.

In addition, LLMs are typically designed to handle dozens of tasks~\cite{lhoest2021datasets}, but these tasks can all be unified under the format of masked language modeling. The reason is that these tasks all involve processing textual data and using the syntax and semantic information within the text. In contrast, GFMs focus a narrower set of tasks but with diverse formats. They excel at graph tasks such as node classification, link prediction and graph classification. The differences in tasks imply that GFMs cannot be learned using methods similar to those in LLMs, significantly increasing the adaptability challenges of GFMs in downstream tasks.

% Tasks include text classification, language generation, question answering, sentiment analysis, etc. 

\subsubsection{Extrinsic Differences}
%Due to the technical limitation on graph foundation models, language foundation models and graph foundation models also extrinsically differ in terms of their underlying architectures, homogenization potential, OOD generalization capabilities, and the types of tasks they are well-suited for. The detailed extrinsic differences are shown as follows:
In addition to the intrinsic differences in data and tasks, there are also some extrinsic differences between GFMs and LLMs, which are due to the lag in technological advancements in GFMs. This section summarizes these differences as follows:

\textbf{Backbone Architectures}. LLMs, such as GPT-3 \cite{brown2020language} and LLaMA \cite{LLaMA}, are mostly based on the Transformer architecture. The advantages of Transformer in terms of expressive power, scalability, parallelizability, and its excellent performance in handling various NLP tasks have made it the mainstream backbone architecture for LLMs. However, for GFMs, using mainstream GNNs as the backbone architecture may not necessarily be suitable. This is mainly because the expressive power and generalization of GNNs have limitations, and their parameter sizes are often too small to exhibit emergent abilities. To enhance the expressiveness of capturing the graph structure, recent works~\cite{sun2022self, sun2024motif, ye2023sincere} efforts to extend GNN to mixed curvature Riemannian space, or to design graph transformers~\cite{sun2021mocl} or models that incorporate LLMs~\cite{TAPE}. However, there is still no unified backbone architecture for GFMs.

\textbf{Homogenization}. LLMs are relatively easy to homogenize. This means that various NLP tasks can be formulated as the same task~\cite{raffel2020exploring}, making it possible to use a single model with unified training paradigm for a wide range of tasks. However, due to the poor transferability of graph structural knowledge, homogenization is more challenging for GFMs. Existing works attempt to achieve homogenization by unifying various tasks as link prediction~\cite{liu2023graphprompt} or graph-level tasks~\cite{allinone}. Additionally, constructing a data-task heterogeneous graph~\cite{PRODIGY} may establish connections between tasks, but it is a more complex process.

\textbf{Domain Generalization}. LLMs have demonstrated strong domain generalization capabilities. They can often perform well on tasks and datasets that were not seen during training, showcasing their ability to generalize across various language-related domains. However, due to the diversity and lack of unified vocabulary of graph data, GFMs generally exhibit weaker generalization across datasets, especially when moving to cross-domain graph data~\cite{zhou2022mentorgnn}. Their performance may degrade significantly when faced with graph structures or characteristics that differ substantially from their training data. Achieving strong domain generalization remains a challenging research problem for GFMs.

\textbf{Emergence}. LLMs have shown emergent abilities, where they can generate coherent and contextually relevant text based on few examples or instructions. Representative emergent abilities include in-context learning~\cite{dong2022survey}, chain of thought reasoning~\cite{wei2022chain} and zero-shot generation~\cite{li2023blip}. However, GFMs have not demonstrated obvious emergent abilities to the same extent as language foundation models. Only a few recent studies discuss the in-context learning~\cite{PRODIGY}, graph reasoning~\cite{NLGraph} and zero-shot graph generation~\cite{su2022molecular} abilities of GFMs.

\subsection{Summary}

In this section, we define the concept of graph foundation models and related technologies, and compares graph foundation models with language foundation models. If readers wish to have a more comprehensive understanding of the concept of GFMs, they can refer to our supplementary materials A and B, where we introduce the impact of graph data and graph tasks on GFMs.
In the following sections, we will introduce three categories of methods for implementing graph foundation models, along with representative works for each method. 
%, as illustrated in Figure~\ref{fig:overall}.
Specifically, GNN-based models use GNN as the backbone architecture, while LLM-based models transform the graph into the input format of LLM and use LLM as the backbone architecture. GNN+LLM-based models, on the other hand, utilize both GNN and LLM as the backbone architecture simultaneously. The distinction in backbone architecture also impacts the methods for pre-training and adaptation. Therefore, in the following sections, we will introduce the backbone architectures, pre-training, and adaptation strategies for each category of methods, seperately.
\section{GNN-based Models}
\label{sec:GM}

Thanks to effective model architectures and training paradigms, language models have achieved remarkable performance in natural language processing tasks. The backbone, pre-training and adaptation techniques employed in language models have inspired a series of corresponding efforts in the field of graph-based tasks. In this section, we will delve into GNN-based models, which draw inspiration from the model architectures or training paradigms used in NLP to apply them to graph tasks. Importantly, unlike the LLM-based models and GNN+LLM-based models to be introduced in the following sections, GNN-based models do not explicitly model text data in their pipeline. 
We have summarized and categorized the works mentioned in this section in supplemental material C.%Table~\ref{tab:gm}.

\subsection{Backbone Architectures}
Numerous GNNs have been proposed and widely used in various graph-related downstream tasks. 
These networks are leveraged for feature extraction, often serving as the foundational components of graph models, commonly referred to as ``backbones''. 
In this subsection, we introduce two advanced GNN backbones: message-passing-based and transformer-based methods.

\begin{figure}[htbp]
    \centering
    \includegraphics[width = 8.5cm]{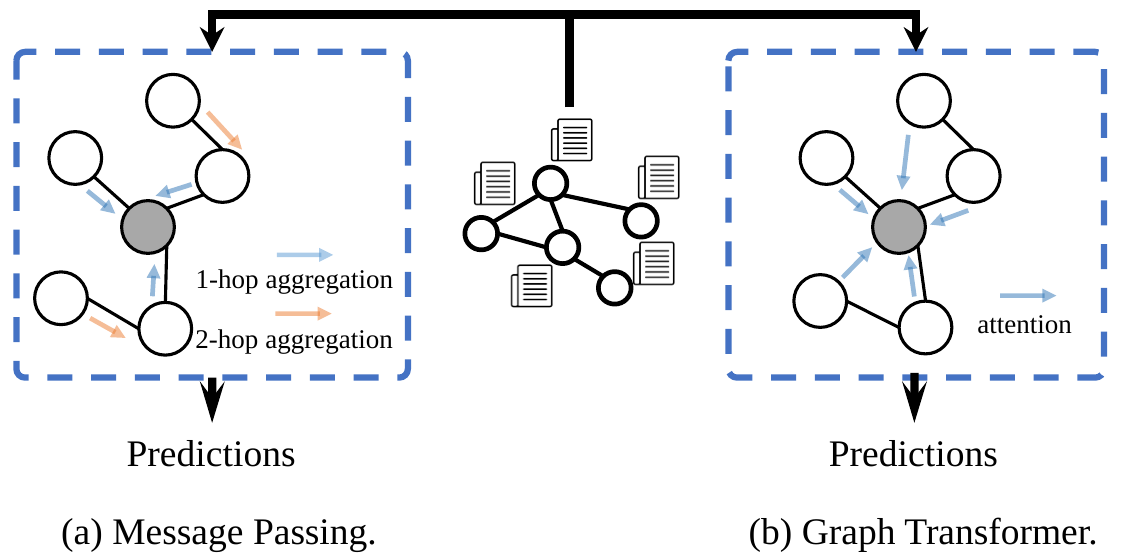}
    \caption{A comparison between message passing-based models and graph transformer. A fundamental distinction is that the message passing mechanism is constrained by the graph structure, and the graph transformer treats the graph as a fully-connected network.}
    \label{fig:backbone}
\end{figure}

\subsubsection{Message Passing-Based Methods}
\label{sec:gm-mpnn}
Message Passing Neural Networks (MPNNs) \cite{gilmer2017neural} represent a broad category of GNN architectures that operate based on the concept of message passing between nodes in a graph. In the message passing mechanism, each node aggregates information from its neighboring nodes, processes the information, and then sends messages to its neighbors in a series of iterative steps.
A typical message passing process can be formulated as :
\begin{equation}
\label{eq:MPNN}
    h_v^{k+1} = U^k(h_v^k, M^k_{u \in N(v)}(h_v^k, h_u^k, \mathbf{X}^e_{(u, v)})), 
\end{equation}
where $h_v^k$ and $h_v^{k+1}$ denote the embedding of node $v$ at layer $k$ and layer $k+1$, $\mathbf{X}^e_{(u, v)}$ denotes the edge attribute of edge $(u, v)$, $N(v)$ denotes the neighbors of node $v$, $M^k_{u \in N(v)}(\cdot)$ and $U^k(\cdot)$ denote the message aggregation and update function at layer $k$.

Many existing GNN-based models utilize message passing-based models as their backbone. Due to the simplicity and effectiveness, several studies~\cite{DGI, GRACE, VGAE, allinone, gong2023ma, PRODIGY} adopt GCN~\cite{kipf2016semi} as their backbone architecture, where GCN~\cite{kipf2016semi} employs a localized first-order approximation of spectral graph convolutions for the dual purpose of capturing graph structure and encoding node features. Several studies~\cite{GraphMAE, GraphMAE2, allinone, PRODIGY} adopt GAT~\cite{velivckovic2017graph} as their backbone architecture, where GAT~\cite{velivckovic2017graph} replaces the average aggregation operation in GCN with a weighted aggregation approach, facilitated by an attention mechanism. Additionally, a multi-head attention technique can be further used in GAT to enhance its performance. GPPT~\cite{sun2022gppt} and VPGNN\cite{wen2023voucher} uses GraphSAGE~\cite{hamilton2017inductive} as their backbone, which operates by sampling a fixed-size subset of neighboring nodes for each target node and then learns embeddings by aggregating and processing these sampled neighbors' embeddings. 
% For heterogeneous graphs, some works ~\cite{GPT-GNN, jiang2021pre, jiang2021contrastive} use HGT~\cite{hu2020heterogeneous} as their backbone, which introduces a specialized attention mechanism for neighborhood aggregation. 
Unlike global attention, HGT employs type-specific parameters to define heterogeneous attention over each edge within the graph. To improve the expressive power, a proportion of studies~\cite{qiu2020gcc, GraphCL, adaptergnn, liu2023graphprompt, guo2023data, fang2022universal} rely on GIN~\cite{xu2018powerful} as their primary architecture, where GIN is a message passing-based model with expressive power theoretically equivalent to a 1-WL test~\cite{weisfeiler1968reduction}. Due to the expressive power of GIN, it is frequently chosen as the backbone for many GNN-based graph models. For an in-depth exploration of message passing-based GNNs, we recommend referring to~\cite{wu2021self, xie2022self, liu2022graph}.
%Due to constraints on the length of this section,

\subsubsection{Graph Transformer-Based Methods}
\label{sec:gm-gt}
% GNNs have shown widespread success for learning on graphs, but they still face fundamental drawbacks, such as limited expressive power, over-smoothing, and over-squashing. Meanwhile, the transformer architecture ~\cite{} has revolutionized natural language processing ~\cite{} and computer vision ~\cite{} tasks, achieving state-of-the-art performance. 
% This success has inspired the design of transformer architectures for graphs ~\cite{} as an appealing alternative to conventional message-passing GNNs. Graph transformers have demonstrated promising results, particularly on molecular prediction tasks ~\cite{}, owning to their fully-connected self-attention mechanism which can overcome the limitations of message-passing GNNs such as over-smoothing and over-squashing ~\cite{}.
While GNNs have demonstrated significant success in learning from graph data, they still confront fundamental limitations, including issues related to limited expressive power~\cite{xu2018powerful}, over-smoothing~\cite{li2018deeper} and over-squashing~\cite{alon2020bottleneck}. In parallel, the transformer architecture ~\cite{vaswani2017attention}, which has revolutionized tasks in natural language processing ~\cite{brown2020language,devlin2018bert} and computer vision ~\cite{dosovitskiy2020image,liu2021swin}, achieving the state-of-the-art performance. It has inspired the development of transformer-based models tailored for graph data ~\cite{ying2021transformers,chen2022structure,kreuzer2021rethinking, kim2022pure}. Graph transformers have exhibited promising results, particularly in molecular prediction tasks ~\cite{ying2021transformers}, owing to their fully-connected self-attention mechanism. This mechanism enables them to address the shortcomings of traditional message-passing GNNs thanks to its long-range modeling capability and strong expressive power. 

The principal distinction between the backbone architectures with message passing mechanism and the graph transformer lies in their treatment of the underlying graph structure. In the case of the graph transformer, it treats the graph as if it were fully connected, meaning it considers and measures the similarity between every pair of nodes in the graph. Conversely, the message passing mechanism operates under the constraint of the adjacency matrix of the graph. It only propagates information between nodes that are explicitly connected in the graph structure. We illustrate the difference between message passing-based models and graph transformers in Figure~\ref{fig:backbone}. 
Currently, there are many research efforts focusing on graph transformers. Here we will present part of these studies that employ a pre-training and fine-tuning learning paradigm: 
% Graph-BERT~\cite{zhang2020graph} takes a novel approach by utilizing sampled linkless subgraphs instead of relying solely on graph links for representation learning. This approach offers more efficient training and mitigates the over-smoothing problem often associated with traditional MPNNs. For pre-training, Graph-BERT suggests two tasks: node classification and graph recovery. By incorporating both tasks into the pre-training phase, the model can capture more comprehensive information from the graph data, resulting in higher performance compared to models trained with a single pre-training task. 
Graph-BERT~\cite{zhang2020graph} uses intimacy-based and hop-based relative positional embeddings to encode node positions in a subgraph. The intimacy-based relative positional embeddings capture the relative positions of nodes in a subgraph based on their connectivity patterns. The hop-based relative distance embeddings capture the relative positions of nodes in a subgraph based on their hop distances. 
% GROVER~\cite{rong2020self} focuses on learning rich structural and semantic information from vast unlabeled molecular data through self-supervised tasks at the node, edge, and graph levels. GROVER integrates message passing networks into a transformer-style architecture, producing more expressive encoders for molecules. 
GROVER~\cite{rong2020self} uses a variant of MPNN called Directed Message Passing Networks (DyMPNs), which can capture the directed nature of molecular graphs and distinguish different types of edges. The DyMPNs in GROVER are used to compute the node and edge embeddings in the Transformer-style architecture.
% MoCL~\cite{sun2021mocl} employs contrastive learning to acquire node or graph representations without the need for labels, enabling it to leverage unlabeled data and reducing the dependency on large amounts of labeled data. Additionally, MoCL incorporates domain knowledge at both local and global levels to enhance representation learning, further improving its overall performance. 
Graphormer~\cite{ying2021transformers} uses spatial encoding to represent node relationships, adding shortest path distance as a bias in softmax attention for better spatial dependency capture. Building upon this foundation, G-Adapter~\cite{G-adapter} introduces a parameter-efficient fine-tuning approach for graph transformers, utilizing Graphormer as its backbone model.
% SimpleDyG~\cite{wu2024feasibility} is a Transformer-based model for dynamic graph modeling that uses a novel temporal alignment technique, simplifying the modeling process without complex modifications.
% Recent works have attempted to extend the Graph Transformer architecture to heterogeneous graphs~\cite{mao2023hinormer,ijcai2024PHGT}.
% Additionally, some studies have delved deeper into the theoretical aspects of Transformer models' performance on graph-structured data. Xing et al.~\cite{xing2024less} identified the over-globalizing problem in Graph Transformers and proposed CoBFormer, a bi-level global graph transformer, to balance local and global information and enhance generalization. Meanwhile, Rosenbluth et al.~\cite{rosenbluth2024distinguished} compared self-attention mechanisms in Graph Transformers with virtual nodes, offering theoretical and empirical analyses to highlight their differences and advantages in handling long-range dependencies.
For a more comprehensive exploration, please refer to other literature on graph transformers~\cite{muller2023attending}.

\subsection{Pre-training}
% 为什么好，意义，
% graph的场景 
% pretrain想解决什么问题 无标签的数据，怎么用起来的问题
Pre-training in the field of NLP involves exposing a model to a vast amount of unlabeled text data, allowing it to learn general language semantic knowledge in a self-supervised manner. This pre-training step equips the model with a foundational understanding of language, enabling it to transfer this knowledge to downstream tasks.
Similarly, the graph domain typically includes many unlabeled nodes and graphs, providing opportunities for pre-training on graphs. Graph pre-training enables the graph models to understand graph structure and semantic information, thus encoding meaningful node or graph embeddings ~\cite{wu2021self, xie2022self}. Recently, some graph pre-training methods have been proposed to learn representations in a self-supervised manner. Based on self-supervised tasks, graph pre-training methods can be categorized into two types: contrastive methods and generative methods.%, as depicted in Figure~\ref{fig:gnn-pretrain}. 
% 解决方法的分类依据，这些工作被分为。。。
% 我们介绍：
% specifically 具体介绍对比
% 总：这类方法的特点，目标（aim）
\subsubsection{Contrastive Methods}
\label{sec:gm-contrastive}
Specifically, the contrastive graph pre-training methods aim to maximize mutual information between different views~\cite{wu2021self}, which forces the model to capture invariant semantic information across various views. The graph view can vary in scale, encompassing local, contextual or global perspectives. These perspectives correspond to node-level, subgraph-level, or graph-level information within the graph, leading to two distinct categories: (1) Same-scale contrastive learning and (2) Cross-scale contrastive learning. Same-scale contrastive learning compares two graph views at the same level. For example, GCC~\cite{qiu2020gcc} uses a node's subgraph embedding as its representation, treating subgraphs of the same node as positives and different nodes as negatives. It applies NCE loss to align positives and separate negatives, capturing general patterns. GraphCL \cite{GraphCL} and GRACE \cite{GRACE} generate two views by graph augmentation and then employ the InfoNCE loss to contrast node-level embeddings, pushing the graph model to acquire the invariant representations. MA-GCL~\cite{gong2023ma} focuses on manipulating the neural architectures of view encoders instead of perturbing graph inputs or model parameters. 
{GCOPE~\cite{zhao2024all} unifies cross-domain graph pre-training using virtual coordinators and contrastive learning, reducing negative transfer and boosting downstream performance. FUG~\cite{zhao2024fug} ensures near-lossless adaptation to diverse graph features with PCA-inspired dimensional encoding and contrastive learning, enabling universal use without preprocessing or model changes.}
% CPT-HG~\cite{jiang2021contrastive} proposes two pre-training tasks. The relation-level pre-training task encodes the relational semantics which constitute the basis of heterogeneity on a heterogeneous graph, while the subgraph-level pre-training task encodes high-order semantic contexts.
% PT-HGNN~\cite{jiang2021pre} considers both node- and schema-level pre-training tasks. The node-level pre-training task utilizes node relations to encourage the GNN to capture heterogeneous semantic properties, while the schema-level pre-training task utilizes the network schema to encourage the GNN to capture heterogeneous structural properties.
Cross-scale contrastive learning compares two graph views at different levels. For example, DGI \cite{DGI} utilizes a discriminator to maximize the mutual information between the node embeddings and the graph embedding and minimize the information between node and corrupted graph embedding.  Such a contrastive process encourages the encoder to capture information of the whole graph. Although DGI enables the model to capture semantic information about nodes and graphs, it ignores the discrepancies between different nodes. 
% To preserve individual information from node-level embeddings, some approaches try to maximize mutual information by contrasting different nodes. 
% Contrastive pre-training methods effectively utilize a large amount of unlabeled graph data by unsupervised contrastive learning, which encourages the graph model to understand the semantic information of graphs. However, such methods cannot generate content that does not exist in the original graph data, which restricts the creativity of models.

\subsubsection{Generative Methods}
\label{sec:gm-predictive}
In addition to contrastive methods, some generative graph pre-training approaches have been proposed. The aim of generative pre-training methods is to enable GNNs to understand the general structural and attribute semantics of graphs. Thus, the GNNs can be adapted to downstream tasks with universal information. 
Generative learning frameworks for graphs can be classified into two categories based on how they acquire generative targets~\cite{xie2022self}: graph reconstruction and property prediction. 

Graph reconstruction aims to reconstruct specific parts of given graphs, emphasizing fidelity in reproducing the original graph structure or node attributes. For example, VGAE~\cite{VGAE} extends the VAE to the graph domain, where it first employs GCN as an encoder to generate node embeddings and then reconstructs the adjacency matrix by the inner product of node embeddings. 
%Although these methods enable models to capture the relationships between nodes and neighbors, they overlook the high-order structural information and attribute semantic information in the graph. 
Furthermore, GPT-GNN \cite{GPT-GNN} proposes the self-supervised edge and attribute generation tasks to push the model to understand the inherent dependencies of attribute and structure. As a result, the model can learn high-order structure and semantic information. GraphMAEs \cite{GraphMAE,GraphMAE2} consider that previous generative methods overemphasize structure information, instead, they employ the reconstruction of features and a re-mask decoding strategy in a self-supervised manner. 
In the property prediction category, models focus on learning and predicting non-trivial properties of provided graphs. For instance, GROVER~\cite{rong2020self} introduces tasks for nodes and edges, predicting context-aware properties within local subgraphs. The graph-level self-supervision task aims to predict motifs, framing it as a multi-label classification problem with each motif as a label.

% GraphGAN \cite{GraphGAN} utilizes the generator and discriminator to force the model to understand the graph data. 
% CG \cite{wan2021contrastive} combines contrastive loss and generative loss to enhance the model to encode both global and local information, resulting in a more meaningful semantic embeddings.
% 总
Although generative methods are capable of generating novel content, the quality and interpretability of the content are hard to guarantee. In the future, it remains to be explored how to enhance the accuracy and rationality of the generative methods.

%\textcolor{red}{In the realm of graph pre-training, recent works have begun to explore pre-training on graph data from multiple domains and then applying the model to graph data in different domains. For example, Davies et al. introduce FOTOM~\cite{davies2023its}, a graph model pre-trained on various graph domains using adversarial contrastive learning, demonstrating its superior performance across multiple downstream tasks compared to single-domain models. Inspired by ControlNet~\cite{zhang2023adding}, GraphControl~\cite{zhu2024graphcontrol} introduces an deployment module to enhance the adaptability of pre-trained models on target datasets by integrating downstream-specific information. These efforts represent a new step forward for graph foundation models.}

\subsection{Adaptation}
\label{sec:adaptation}
% % 一小段帽子 总起
% 背景（和pretrain有关系
% 核心问题（怎么adaptation，
% solution：两种方式（graph 变模型参数，变输入？？check
Typically, the objectives of pre-training tasks are different from the downstream ones, which hinders the transferability of pre-training models. To this end, fine-tuning is a common adaptation approach based on subtle adjustments of model parameters. In addition, the \emph{``pre-train, prompt and predict''} paradigm has attracted considerable attention in recent years. By using prompts, the format of downstream tasks is aligned with that of pre-training tasks, enabling pre-training models to handle downstream tasks in a more effective manner.
%To fully utilize the semantic information learned by the pre-trained models, several measures have been proposed to guide the adaptation of the pre-training model to downstream tasks.to adapt pre-trained models for downstream tasks 

\subsubsection{Fine-Tuning}
\label{sec:gm-ft}
% 背景：什么场景用
% 数据/任务迁移
% 传统model head tune
% 具体介绍 model head tune
For the situation where the model conducts the pre-training and downstream tasks in the same domain, we can utilize a pre-training model to generate node embeddings or graph embeddings, and subsequently fine-tune an external task-specific layer to generalize the pre-training model to downstream tasks. DGI \cite{DGI} and GRACE \cite{GRACE} utilize the pre-trained encoder to obtain node embeddings, and then fine-tune a logistic regression classifier with labeled data to handle the node classification task. 
Additionally, there is a more practical scenario where pre-training is performed on the known graphs while tested on unseen graphs. Pre-training models cannot encode unknown graphs appropriately, thus we need to fine-tune the model in this situation. GPT-GNN \cite{GPT-GNN} employs the labeled data to fine-tune a downstream task-specific decoder, which guides the pre-training model to adapt to downstream tasks. 
Moreover, some parameter-efficient fine-tuning methods have been proposed recently. AdapterGNN \cite{adaptergnn} employs two parallel adapters before and after the message passing stage to modify the input graph. Such addition-based methods only need to fine-tune the introduced parameters. G-Adapter \cite{G-adapter} proposes a parameter-efficient fine-tuning method for graph transformer, which introduces graph structure into the fine-tuning by message passing. G-TUNING~\cite{sun2024fine} is a fine-tuning strategy for GNNs that utilizes graphon reconstruction to preserve generative patterns and address structural divergence between pre-training and downstream datasets

Although the fine-tuning methods have achieved significant success, they typically require sufficient labeled data to tune the model parameters. Moreover, conventional fine-tuning methods necessitate repetitive fine-tuning for each task, incurring significant computational costs.
Therefore, more advanced fine-tuning techniques for graph foundation models are still to be explored.
% 进一步，效率
% Parameter-efficient Fine Tuning
% delta tuning
% 评价：fine tune缺点（为prompt 铺垫

\subsubsection{Prompt Tuning}
\label{sec:gm-pt}
% 背景：什么场景用 tune不动 样本少的时候，引出prompt tuning。
% solution：统一（我们想探究的核心问题），设计
% challenge：node graph link。。。。
% 具体：
% 节点分类+链接预测
% node level， graph level
% 。。。jure 进一步 task graph。。。
% 。。。
% 总结（看future works。。。。
% Due to the inconsistency between the pre-training task and the downstream task, the fine-tuning method may not effectively adapt the pre-trained model to the downstream task. 

% Recently, prompt tuning is proposed to narrow the gap between the downstream and pre-training task, consequently maximizing the potential of the pre-trained model to attain generalization in the downstream task, which has garnered remarkable achievements in the field of NLP. 
Prompt tuning has recently emerged as a strategy to circumvent the need for full-parameter tuning, {facilitating both multi-task adaptation and zero-shot learning~\cite{liu2023pre,lester-etal-2021-power,liu-etal-2022-p}. This innovative approach has found significant applications in graph data, where recent studies have concentrated on using prompt tuning to enhance the performance and adaptability of GNN-based models.
Following the framework proposed by Guo et al.~\cite{guo2023data}, these methods can be categorized into two distinct groups: pre-prompt methods and post-prompt methods, based on whether the task-specific prompts operate before or after the backbone module. } 

\textbf{Pre-prompt methods} {modify the input graph's topology or node features before message passing to aid downstream tasks or construct a prompt graph to enhance model adaptation.} For example, AAGOD~\cite{guo2023data} proposes to implement the data-centric manipulation by superimposing an amplifier on the adjacency matrix of the original input graph as learnable instance-specific prompts. All In One \cite{allinone} converts the node-level and graph-level tasks to graph-level tasks. It treats an extra subgraph as a prompt and merges it with the node subgraph. The model subsequently utilizes the combined graph to generate predictions. GPF~\cite{fang2022universal} introduces a uniform feature modification vector to each node in the graph, which can be optimized to adapt pre-training GNN models under any pre-training strategy. 
%SGL-PT~\cite{zhu2023sgl} incorporates a strong and general pre-training task that harnesses the advantages of both generative and contrastive approaches. 
Additionally, it features verbalizer-free prompting function, thus aligning the downstream task with the pre-training method's format. PRODIGY~\cite{PRODIGY} is a framework for pre-training an in-context learner over prompt graphs. The goal is to enable a pretrained model to adapt to diverse downstream tasks without optimizing any parameters. IGAP~\cite{yan2024inductive} bridges the gap between graph pre-training and inductive fine-tuning by addressing the graph signal and structure gaps using learnable prompts in the spectral space. {TPP~\cite{niu2024replay} achieves a replay-free and forget-free GCIL system by storing task-specific knowledge in compact learnable tokens using graph prompts with a frozen pre-trained GNN, avoiding model updates or data replay.}

\textbf{Post-prompt methods} {use task-specific prompts on representations after message passing to aid downstream task adaptation.}
For example, GPPT \cite{sun2022gppt} employs a prompting function to generate a token pair for each class, transforming all node classification tasks into link prediction tasks. GraphPrompt \cite{liu2023graphprompt} unifies pre-training and downstream tasks into a consistent task format based on subgraph similarity, and utilizes labeled data to learn a task-specific prompt vector for each task, which modifies the model's Readout operation and narrows the gap between link prediction and downstream tasks. 
GraphPrompt+~\cite{yu2023generalized} further enhances GraphPrompt by generalizing pre-training tasks and employing layer-wise prompts to capture hierarchical knowledge across the graph encoder, improving task-specific knowledge extraction for both node and graph classification. {ProNoG~\cite{pronog} uses conditional prompting for non-homophilic graphs, leveraging a condition-net to generate node-specific prompts that refine embeddings for fine-grained task adaptation.}

% Notably, there are exceptions that incorporate both pre-prompt and post-prompt methods. For example, MultiGPrompt \cite{yu2024multigprompt} aims to unify the pre-training and downstream tasks across both homogeneous and heterogeneous graphs using a dual-template design. HGPROMPT~\cite{yu2024hgprompt} presents a novel few-shot prompt learning framework that bridges the gap between homogeneous and heterogeneous graphs by using a dual-template design for pre-training and downstream tasks.

Although these methods have improved the performance in few-shot scenarios, further exploration is needed to understand the semantics and interpretability of the graph prompts.

% The prompt technique has recently attracted significant attention in the NLP field, which does not tune the based model or external part to adapt to downstream tasks. Instead, prompts bridge the gap between the pre-train task and the downstream task. Typically, prompts can be classified as discrete prompt and continuous prompts. In the field of NLP, discrete prompts are typically concat directly to the original text with explicit words. In the graph, such discrete prompts are difficult to define and poorly understood, which leads to the challenge of designing discrete prompts in graphs. \cite{allinone} treats an extra graph as a prompt and concatenates it to the original node subgraph. Furthermore, some methods utilize a small amount of labeled data to supervise the optimization of prompts, which learns a soft prompt. GPPT \cite{sun2022gppt} employs a prompting function to generate a token pair for each class, transforming all node classification tasks into link prediction tasks. GraphPrompt \cite{liu2023graphprompt} considers that the ReadOUT operation differs for different downstream tasks. It utilizes labeled data to learn a task-specific prompt vector for each task, which modifies the model's ReadOut operation and transfer the model to new tasks.

\subsection{Discussion}

GNN-based models offer several advantages, particularly their {ingenious inductive bias and} compact parameter size. These models naturally possess essential properties like permutation invariance, enabling them to handle graph-structured data effectively. Additionally, GNN-based models offer the advantage of low-cost training and efficient resource usage, which reduces computational requirements and makes them accessible for deployment even in resource-constrained environments. Moreover, they can generalize well from small amounts of labeled data. By propagating label information through the graph, they can enhance prediction accuracy even when labeled data is sparse. 

{Furthermore, for more complex graph data such as heterogeneous graphs, hypergraphs, and temporal graphs, preliminary research have investigated GNN-based graph foundation models. For example, CPT-HG~\cite{jiang2021contrastive} and PT-HGNN~\cite{jiang2021pre} have designed sophisticated pre-training methods tailored for high-order semantic information in heterogeneous graphs. MultiGPrompt~\cite{yu2024multigprompt} and HGPROMPT~\cite{yu2024hgprompt} both use prompt-based learning to link pre-training and downstream tasks on heterogeneous graphs. PhyGCN~\cite{deng2024pretrained} and IHP~\cite{yang2024instruction} use self-supervised hyperedge prediction and instruction-based prompts respectively to improve node representations in hypergraphs. GraphST~\cite{zhang2023spatial} and GPT-ST~\cite{li2023generative} use pre-training to improve spatio-temporal representations for temporal graphs.}

% contexts such as hypergraphs~\cite{yang2024instruction, deng2024pretrained}, heterogeneous graphs~\cite{yu2024hgprompt, jiang2021pre, ma2024hetgpt}, and temporal graphs~\cite{li2023generative, zhang2023spatial}. 

Despite their many advantages, GNN-based models have several notable disadvantages. One primary limitation is their lack of explicit text modeling. They often underutilize the rich semantic information embedded in textual attributes associated with nodes or edges, leading to suboptimal exploitation of textual data. {Another significant drawback is the limited capacity of GNN-based models to incorporate and utilize general knowledge effectively. Unlike LLMs, which are pre-trained on vast corpora of text and can leverage extensive world knowledge, GNN-based models typically lack such pre-training on diverse and comprehensive datasets. This gap restricts their ability to generalize across different domains and limits their performance in tasks requiring broad contextual understanding or common-sense reasoning.}

{One promising direction is the integration of LLMs with GNN-based models. LLMs can provide a robust framework for incorporating extensive textual knowledge, enhancing the models' ability to understand and utilize semantic information embedded in text.} In the following sections, we will explore graph learning models that incorporate LLMs.
\section{LLM-based Models}
\label{sec:LLM}
% 介绍+定义:graph+text
% where powerful?
% why textual?
Researchers are actively exploring ways to leverage LLMs as core and sole backbone for graph learning~\cite{InstructGLM, GPT4Graph, NLGraph}, for the following advantages that can not be underestimated. Firstly, transformer-based models show a remarkable ability to seamlessly integrate textual information in graph data~\cite{InstructGLM}. Additionally, employing a LLM-liked backbone empowers models to unify diverse graph learning tasks, as these tasks can be described using natural language. Furthermore, recent advancements, such as NLGraph \cite{NLGraph}, GPT4Graph \cite{GPT4Graph}, showcase the LLMs' prowess in preliminary graph reasoning. These advantages mark a highly promising direction for the development of such models. To discover the potential of engaging LLMs into graph learning, these works involve both graph-based properties and textual information as input for the backbone networks.
% 定义LLM以及原因
% following xxx KG+LLM，exploring， not narrowly...
Following some surveys \cite{pan2023unifying,chen2023exploring}, our characterization of the backbone is not confined solely to the narrow definition of LLMs (like GPT-4); it also encompasses certain transformer-based models that leverage textual information. We have summarized and categorized the works mentioned in this section in supplemental material D.%Table ~\ref{tab:llm}.

\begin{figure*}[htbp]
\centering
\subfigure[Graph-to-token.]{
\parbox[][4.5cm][c]{0.45\linewidth}{
\begin{minipage}[t]{\linewidth}
\centering
\includegraphics[width=\linewidth]{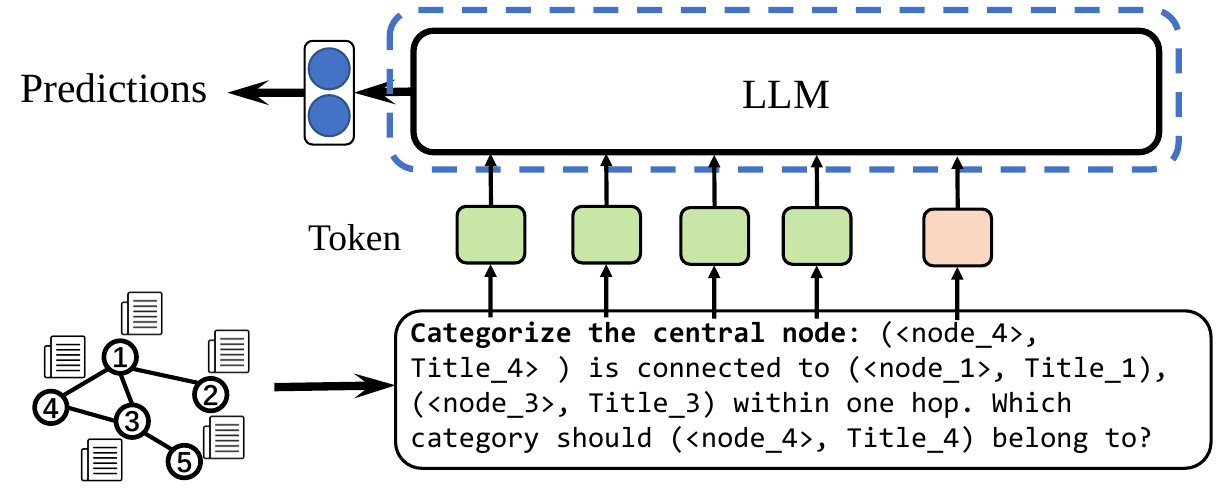}
\end{minipage}
}}
\hspace{0cm} % 调整这里的间距
% \noindent
\subfigure[Graph-to-text.]{
\parbox[][4.5cm][c]{0.45\linewidth}{
\begin{minipage}[t]{\linewidth}
\centering
\includegraphics[width=\linewidth]{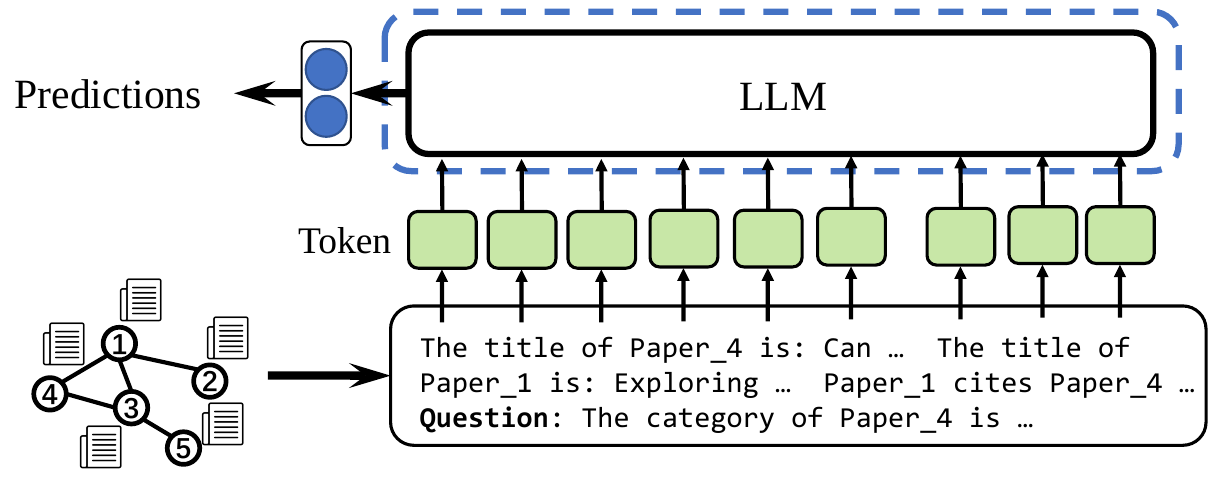}
\end{minipage}
}}%     
\caption{An illustration of two existing approaches to align graph data with natural language. One approach tokenizes graph data and use node representations (depicted as red tokens) as well as text tokens (depicted as green tokens) to be the input of LLMs. Another approach represents graph data with prompts in natural language and uses text tokens only (depicted as green tokens) as the input of LLMs.}
\label{fig:StandaloneLLMs}
\end{figure*}

% \begin{figure*}[t]
% \centering
% \subfigure[Graph to token.]{
% \begin{minipage}[t]{1\linewidth}
% \centering
% \includegraphics[width=0.9\linewidth]{images/sec5_1.1.pdf}
% \label{fig:graph2token}
% \end{minipage}%
% }%

% \subfigure[Graph to text.]{
% \begin{minipage}[t]{1\linewidth}
% \centering
% \includegraphics[width=0.9\linewidth]{images/sec5_2.1.pdf}
% \label{fig:graph2text}
% \end{minipage}%
% }%

% \centering
% \vspace{-3mm}
% \caption{An illustration of two existing appoaches to align graph data with natural language. One approach represents graph data with prompts in natural language. Another approach tokenizes graph data and use node representations(depicted as red tokens in the figure) as well as text tokens(depicted as green tokens in the figure) to be the input of LLMs.}
% \label{fig:StandaloneLLMs}
% \end{figure*}

%While using LLMs in specific graph tasks encounters challenges like imprecise inference, lack of suitable evaluation methods and significant time costs, their advantages should not be underestimated. Firstly, transformer-based models demonstrate a remarkable ability to seamlessly integrate multimodal information, enhancing perception across various modality tasks. Additionally, employing a LLM-liked backbone empowers models to unify diverse graph learning tasks, as these tasks can be described using natural language. Furthermore, recent advancements, such as NLGraph \cite{NLGraph}, GPT4Graph \cite{GPT4Graph}, showcase the LLMs' prowess in preliminary graph reasoning, marking a highly promising direction for the development of such models.

% \input{tables/llm-table}

\subsection{Backbone Architectures}
% LLM+graph的核心问题（align）
A central question in employing LLMs for graph data is how to align graph data with natural language so that LLMs can understand them. 
% 解释一下：压成token以后无structure信息，text；解释一下什么是token？
% LLM有两种对齐的点，一个是token一个是text，介绍一下token。对齐的方式有两种，一种是graph-》token(transformer的输入）；另一种是graph->文本->token。直接转成token和间接转成token。
Given that LLMs initially accept tokens as their inputs and rely on self-attention layers to process input sequences for producing hidden representations, it can be a difficult task to attain a fine-grained modeling of graph structure information~\cite{InstructGLM}. %加引用
% As illustrated in Figure~\ref{fig:StandaloneLLMs}, there are primarily two approaches that have been developed to address this crucial question, namely graph-to-token and graph-to-text.
{As illustrated in Figure~\ref{fig:StandaloneLLMs}, we categorize existing LLM-based methods into two types, graph-to-token and graph-to-text.}
% The first type, graph-to-text methods, directly incorporate graph data into LLMs by converting the graph structure into a textual format. This textual representation, combined with task-specific instructions, is then used to query the LLM. The second type, graph-to-token methods, encode real-world graphs by condensing their complex structures and lengthy descriptions into token-level embeddings, referred to as Graph Tokens. These Graph Tokens are appended to natural language instruction tokens to query the LLM.
{The key distinction between these two approaches lies in the use of an additional encoder. Graph-to-token methods rely on an additional encoder (e.g., BERT) to generate embedding-level representations for each node, while graph-to-text method directly translates graph representations into natural language input for LLMs without the need for an additional encoder.} 
% Furthermore, graph-to-token methods require updating the LLM’s vocabulary, necessitating the use of open-source, trainable LLMs like LLaMA, whereas graph-to-text methods can be applied to proprietary LLMs such as ChatGPT. Graph-to-text methods typically require no additional training and are more computationally efficient. However, they are constrained by the LLM’s input length, limiting the size of the graph data they can handle. In contrast, graph-to-token methods incur higher computational costs but can process large-scale graph data encountered in real-world scenarios, as each node can usually be represented by just a single token~\cite{InstructGLM}.

\subsubsection{Graph-to-token}
\label{sec:llm-token}
One approach entails the tokenization of graph information and imitates the standardized input format of transformer-based models. This methodology necessitates not only the serialization of graph data into tokens but also the solutions for encoding the graph's structural information. 
% 不同的backbone需求，开源的LLMs or transformers only
Since this method uses node representations as unique tokens for the input of backbone models, the backbone need to be either trainable transformers or open source LLMs. For instance, InstructGLM \cite{InstructGLM} uses LLaMA \cite{LLaMA} and T5 \cite{raffel2020exploring} to be their backbones for further tuning.

The concept of graph-to-token initially surfaces in GIMLET \cite{zhao2023gimlet} that treats node representations as tokens and aims to integrate graph data with textual data. Specifically, GIMLET expands the capabilities of LLMs to accommodate both graph and text data by using the transformer architecture, incorporating generalized position embedding and instruction-based pre-training. Furthermore, efforts have been made to integrate graph data with other modalities beyond just text data. For instance, Meta-Transformer \cite{zhang2023meta} introduces a transformer-based architecture designed to incorporate various forms of multimodal data, including graphs, text, and images. However, despite the promising trend indicated by developing unified multimodal intelligence using a transformer backbone, their approaches cannot be considered as graph foundation models because they do not involve any pre-training and adaptation learning paradigm.

InstructGLM \cite{InstructGLM} on the other hand, adopts a pre-training and adaptation framework and introduces LLMs to further enhance the model's text processing capabilities, making it a strong contender for the position of a graph foundation model. In this framework, the vocabulary of the LLMs is expanded by incorporating the inherent node feature vectors from the graph as distinct and unique tokens for each node. Leveraging LLMs and the transferability of natural language, InstructGLM makes a valuable contribution to the ongoing movement towards graph foundation model architectures and pipelines that span multiple modalities.

These efforts tokenize graph data to align it with natural language, enabling joint understanding with data from other modalities. Their conclusions showcase promising results for integrating graph data with natural language. However, despite these promising results, how to inform LLMs of underlying graph structures remains an important challenge in this approach.
\subsubsection{Graph-to-text}
\label{sec:llm-text}
To align graph data with natural language, another approach involves describing graph information using natural language. Several approaches have been developed along this line of thoughts, utilizing prompts to integrate the capabilities of LLMs into classical tasks on graphs. For this method, which exclusively relies on natural language prompts, the associated backbone model can be any LLM, even if it is not open-sourced. 
For instance, Graph-LLM \cite{chen2023exploring} utilizes multiple language models of different sizes, including BERT \cite{devlin2018bert}, DeBERTa \cite{he2020deberta}, Sentence-BERT \cite{reimers2019sentence}, GPT-4 \cite{OpenAI_2023} and LLaMA \cite{LLaMA}.
% For instance, LLMtoGraph \cite{liu2023evaluating} uses several LLMs including GPT-4 \cite{OpenAI_2023} and Vicuna \cite{chiang2023vicuna}, TextForGraph \cite{wenkel2023pretrained} uses GPT-2 and GPT-3 \cite{OpenAI_2023}, When\&Why \cite{huang2023can} uses ChatGPT APIs \cite{OpenAI_2023} and LLaMA \cite{LLaMA}, GraphWiz \cite{chen2024graphwiz} uses LLaMA \cite{LLaMA} and Mistral \cite{jiang2023mistral}, CGForLLM \cite{antonucci2023zero} uses GPT-4 \cite{OpenAI_2023}, LLM4DYG \cite{zhang2023llm4dyg} uses LLaMA \cite{LLaMA}, Vicuna \cite{chiang2023vicuna} and GPT-3.5 \cite{OpenAI_2023}, GPT4Graph \cite{GPT4Graph} uses InstructGPT-3 \cite{ouyang2022training}, LLM4Mol \cite{qian2023can} uses GPT-4 \cite{OpenAI_2023}, Graph-LLM \cite{chen2023exploring} utilizes multiple language models of different sizes, including BERT \cite{devlin2018bert}, DeBERTa \cite{he2020deberta}, Sentence-BERT \cite{reimers2019sentence}, GPT-4 \cite{OpenAI_2023} and LLaMA \cite{LLaMA}.

Initial attempts mainly use edge list to describe graph structures in natural language and make assessment on various graph tasks. NLGraph \cite{NLGraph} also conducts a comprehensive assessment of LLMs across eight graph reasoning tasks as well as popular GNN tasks in natural language. Similarly, utilizing edge lists to describe graph structure, the results once again underscores the limitations of this approach when dealing with complex graph problems. TextForGraph \cite{wenkel2023pretrained} designed two types of prompts, full text and reduced text, to describe information on the graph, effectively compressing the prompt length. When\&Why \cite{huang2023can} designs several styles of prompts and offers key insights into the use of LLMs for processing structured data. GraphWiz \cite{chen2024graphwiz} designs different prompts for various tasks on graphs, including cycle detection, subgraph matching, and more.

Moreover, GPT4Graph \cite{GPT4Graph} introduces a novel approach to prompt engineering that combines manually crafted prompts with prompts generated by the language model itself, referred to as handcrafted prompts and automatic prompts. Specifically, for manual prompting, it utilizes description languages such as edge lists, adjacency lists, Graph Modeling Language (GML), and Graph Markup Language (GraphML) to represent graph structures and compare their effectiveness. For automatic prompting, it employs techniques like graph summarization, neighborhood summarization, graph exploration, and graph completion to actively engage LLMs in understanding and manipulating graphs, facilitating graph-based reasoning and learning. The findings indicate that self-prompting is a more effective method for informing LLMs about graph structures. Graph-LLM \cite{chen2023exploring} further supports this conclusion, emphasizing that neighbor summarization is the most effective technique in existing prompt engineering methods.

% Furthermore, efforts have been undertaken to explore the potential of LLMs in addressing tasks from other domains using textual prompts. For instance, LLM4Mol \cite{qian2023can} uses simplified molecular input line entry system (SMILES) to directly describe the property of molecule, and explores how LLMs can contribute to molecular property prediction.

These studies highlight significant potential for using natural language to describe graph data and using textual tokens as the input of LLMs for graph learning. Nevertheless, a key takeaway from their conclusions is that, at the present moment, the way we use these prompts may not be an effective approach for mining underlying graph structures.

% These studies highlight significant potential for using natural language prompts in graph-related tasks during adaptation stage. Nevertheless, a key takeaway from their conclusions is that, as of now, using prompts may not be an effective approach for mining graph structures.
\subsection{Pre-Training}
% 说明这一节介绍LLM的pretrain方法
The methods discussed in this section solely employ LLMs as the backbone. Hence, the pre-training phase for these methods corresponds to the pre-training phase of LLMs. There are mainly two tasks used in LLM-based models for graph learning, we will provide a concise overview of these two pre-training tasks.

\vspace{-2mm}
\subsubsection{Language Modeling (LM)}
\label{sec:llm-lm}
% 哪些方法配套地使用了哪种预训练
Language Modeling (LM) is one of the most common self-supervised task in NLP, and is widely adopted by many state-of-the-art LLMs, such as LLaMA \cite{LLaMA} and GPT-3 \cite{ouyang2022training}. LM task can be essentially addressed to the challenge of estimating probability distributions of the next word. While LM represents a broad concept, it frequently pertains specifically to auto-regressive LM or unidirectional LM in practical applications~\cite{qiu2020pre}. Many methods involve LM as their pre-training method, namely InstructGLM \cite{InstructGLM}, NLGraph \cite{NLGraph}, GPT4Graph \cite{GPT4Graph}, Graph-LLM \cite{chen2023exploring}, TextForGraph \cite{wenkel2023pretrained}, When\&Why \cite{huang2023can}, GraphWiz \cite{chen2024graphwiz} and CGForLLM \cite{antonucci2023zero}.
%Due to their choices of backbone networks, all approaches mentioned in this chapter have

In the context of a text sequence represented as $s_{1:\mathit{L}}= [\mathit{s}_{1},\mathit{s}_{2},\cdots,\mathit{s}_{\mathit{L}}]$, its overall joint probability, denoted as $\mathit{p}\left ( s_{1:\mathit{L}} \right )$, can be expressed as a product of conditional probabilities, as shown in equation:
\begin{equation}
    \mathit{p}\left ( s_{1:\mathit{L}} \right ) = \prod_{\mathit{l} = 1}^{\mathit{L}}\mathit{p}\left ( \mathit{s}_{\mathit{l}}|s_{0:\mathit{l}-1} \right ).
\end{equation}
Here, $s_{0}$ represents a distinct token signifying the commencement of the sequence. The conditional probability $\mathit{p}\left ( \mathit{s}_{\mathit{l}}|s_{0:\mathit{l}-1} \right )$ is essentially a probability distribution over the vocabulary based on the linguistic context $s_{0:\mathit{l}-1}$. To model the context $s_{0:\mathit{l}-1}$, a neural encoder $\mathit{f}_{nenc}(\cdot )$ is employed, and the conditional probability is calculated as follows:
\begin{equation}
    \mathit{p}\left ( \mathit{s}_{\mathit{l}}|s_{0:\mathit{l}-1} \right ) = \mathit{f}_{lm}(\mathit{f}_{nenc}(s_{0:\mathit{l}-1})).
\end{equation}
In this equation, $\mathit{f}_{lm}$ represents the prediction layer. 
% 优缺点
By training the network using maximum likelihood estimation (MLE) with a large corpus, we can effectively learn these probabilities. 
% However, unidirectional language models have a limitation in that they encode contextual information for each token based solely on the preceding leftward context tokens and the token itself, but an improved contextual representation of text should ideally capture contextual information from both forward and backward directions.
Nevertheless, a drawback of unidirectional language models is their encoding of contextual information for each token, which is solely based on the preceding leftward context tokens and the token itself. However, for more robust contextual representations of text, it is preferable to capture contextual information from both the forward and backward directions.
\subsubsection{Masked Language Modeling (MLM)}
\label{sec:llm-mlm}
Masked language modeling (MLM) is introduced to address the limitation of the traditional unidirectional language model, frequently denoted as a Cloze task \cite{qiu2020pre}. In MLM, specific tokens within the input sentences are randomly masked, and the model is then tasked with predicting these masked tokens by analyzing the contextual information in the surrounding text. As an effective pre-training task, MLM is adopted in BERT \cite{devlin2018bert} and T5 \cite{raffel2020exploring}. Additionally, MLM can be categorized into two subtypes: Sequence-to-Sequence MLM (Seq2Seq MLM) and Enhanced MLM (E-MLM). Seq2Seq MLM involves utilizing the masked sequences as input for a neural encoder, and the resulting output embeddings are used to predict the masked token through a softmax classifier. On the other hand, E-MLM extends the mask prediction task to various types of language modeling tasks or enhances MLM by integrating external knowledge. MLM also faces some drawbacks as this pre-training method may result in a disconnection between the pre-training and fine-tuning stages since the mask token is absent during fine-tuning. InstructGLM \cite{InstructGLM} and Graph-LLM \cite{chen2023exploring} use T5/BERT as backbones, and adopt MLM pre-training strategy.%, because they choice of backbones include T5 and BERT.

% 其他的预训练方法，并说明未来的可能性
There are also many pre-training tasks like Permuted Language Modeling (PLM) \cite{yang2019xlnet}, Denoising Autoencoder (DAE) \cite{lewis2019bart}, Text-Text Contrastive Learning (TTCL), \cite{arora2019theoretical} and others. These pre-training tasks are currently not adopted by existing LLM-based models in graph learning, and thus not within the scope of our discussion in this section. However, we believe that in the future, more research will be developed on graph tasks involving these pre-training tasks, offering additional possibilities for the establishment and refinement of graph foundation models.

\subsection{Adaptation}
The adaptation phase plays a pivotal role in enhancing the performance of LLM-based models in graph learning. LLMs are primarily trained on textual corpora, which results in a significant gap between the pre-training phase and their deployment on graph tasks. Both the graph-to-token and graph-to-text methods are accompanied by specific adaptation techniques designed to enhance the LLM's ability to understand graph data effectively. As these methods share a fundamentally similar training procedure that utilizes prompts, we classify these adaptation strategies in the aspect of prompt engineering: manual and automatic. %In the following sections, we will further introduce the specific adaptation strategies utilized by these methods.
\subsubsection{Manual Prompting}
\label{sec:llm-manual}
%InstructGLM \cite{InstructGLM} involves a method known as instruction prompting to represent graph structures using natural language. Their approach designs a set of graph descriptions centered around a central node, accompanies by task-specific descriptions. Through instruction prompting, InstructGLM aims to enhance the understanding of LLMs while empowering neural models with the capability to effectively tackle a variety of tasks by incorporating natural language instructions. Instruction prompting presents another promising avenue for integration with significant development potential.
Methods mentioned here use manually created prefix style prompts. For instance, LLMtoGraph \cite{liu2023evaluating} and NLGraph \cite{NLGraph} employ node and edge lists incorporating other graph properties described in natural language to form a comprehensive prompt. GPT4Graph \cite{GPT4Graph} goes a step further by utilizing additional description languages to represent graph data, such as edge list, adjacency list, GML and GraphML, providing a more extensible framework for manual graph prompts. Furthermore, InstructGLM \cite{InstructGLM} employs instruction prompting to involve the design of a set of graph descriptions centered around a central node, coupled with task-specific descriptions.
Graph-LLM \cite{chen2023exploring}, TextForGraph \cite{wenkel2023pretrained}, When\&Why \cite{huang2023can}, GraphWiz \cite{chen2024graphwiz} and CGForLLM \cite{antonucci2023zero} also use natural language instructions and subsequently conduct a series of comprehensive experiments.
\subsubsection{Automatic Prompting}
\label{sec:llm-automatic}
Creating prompts manually is a time-consuming task, and these prompts can sometimes be sub-optimal \cite{jiang2020can}. To address these drawbacks, automatically generated prompts have been introduced for further adaptation. GPT4Graph \cite{GPT4Graph} firstly tries to employ three different types of prompts generated by LLM itself, namely graph summarization, graph exploration and graph completion, in graph tasks. Specifically, graph summarization generates a summary of the given graph by extracting key features or a summary of the neighborhood information of target nodes. Graph exploration means generating a sequence of queries or actions to retrieve information from the given graph. Graph completion generates partial graphs and prompt itself to complete the missing parts. By leveraging these self-prompting strategies, LLMs can actively engage in the understanding and manipulation of graphs, facilitating graph-based reasoning and learning. Graph-LLM \cite{chen2023exploring} uses automatic prompts as well in the form of neighbor summary, and their experimental results once again emphasize the efficiency of automatic prompting.
% Creating prompts manually is a time-consuming task, and these manually crafted prompts can occasionally fall short in terms of optimization \cite{jiang2020can}. To overcome these limitations, the adoption of automatically generated prompts has gained attention. In the context of graph-related tasks, GPT4Graph \cite{GPT4Graph} introduced three distinct self-prompting strategies generated by the LLM itself: Graph summarization, Graph exploration, and Graph completion. Graph summarization involves the generation of summaries for given graphs, extracting key features or neighborhood information around target nodes. Graph exploration entails generating a sequence of queries or actions to extract information from the provided graph. Graph completion make the LLM to create partial graphs and then task itself with completing the missing components. By incorporating these self-prompting mechanisms, LLMs actively participate in comprehending and manipulating graphs, thereby enhancing graph-based reasoning and learning. Graph-LLM \cite{chen2023exploring} utilizes automatic prompts as well, particularly in the form of neighbor summarization. The experimental results from their work underscore the efficiency of automatic prompts.

% 说明一些其他的adaptation方法（fine-tuning类）并说明未来可能性
Additionally, there are various adaptation approaches based on fine-tuning, including Vanilla Fine-Tuning \cite{devlin2018bert}, Intermediate Fine-Tuning (IFT) \cite{poth2021pre}, Multi-task Fine-Tuning (MTFT) \cite{liu2019multi}, and Parameter Efficient Fine-Tuning \cite{houlsby2019parameter}. These methods offer efficient ways to adapt pre-trained models to downstream tasks, although they have not been applied to graph tasks at this time. However, we anticipate that future research will explore the integration of these adaptation approaches into graph tasks, further advancing the development of the graph foundation model.
\subsection{Discussion}
%结论，加优缺点，写本章的总结
Efforts of aligning graph data with natural language and using sole LLMs as graph learners has paved the way for exciting developments. The integration of graph data, text, and other modalities into transformer-based models presents a promising way, with the potential to connect techniques from the GNN field with advancements in the LLM domain. Moreover, leveraging LLMs allows for the unification of various graph tasks, as these tasks can all be described in natural language. This makes LLM-based backbones a more competitive selection for building GFMs.

%缺点: 无法理解图、建模图，非文本图处理、不好微调？
Nonetheless, it is essential to acknowledge that the current ways of utilizing LLMs as backbones for graph learning also presents inherent limitations. These limitations encompass the inability of LLMs to {effectively process the lengthy textual information required to describe graph structures}, their incapacity to engage in multi-hop logical reasoning through graph links, the challenge they face in capturing the topological structures prevalent in highly connected graphs, and their struggle in handling the dynamic nature of graphs that evolve over time. {Furthermore, graph-to-text methods are constrained by the LLM’s input length, limiting the size of the graph data they can handle. In contrast, graph-to-token methods incur higher computational costs but can process large-scale graph data encountered in real-world scenarios, as each node can usually be represented by just a single token~\cite{InstructGLM}.} These shortcomings underscore the need for further research in using LLM-based models for graph learning. 

% 未来研究方向
{Future research directions for LLM-based approaches include enhancing the ability of LLMs to more effectively and efficiently understand critical information in graphs, including node features and topological structures. Considering that LLMs cannot directly comprehend graphs, and flattened natural language description of graphs are likely to result in information loss, efficient and structured modeling techniques for graphs need to be developed. These methods are expected to help bridge the gap between natural language prompts and the comprehensive information present in graph data. {Moreover, existing research, such as LLM4DYG~\cite{zhang2023llm4dyg}, has explored the application of LLMs to complex graph data, specifically temporal graphs. However, more diverse types of graph data, such as hypergraphs and heterogeneous graphs, need to be explored.}}
\section{GNN+LLM-based Models}
\label{sec:LLM+GM}
\begin{figure*}[htbp]

\centering
\hspace{0cm}
\subfigure[GNN-centric methods.]{
\parbox[][4cm][c]{0.28\linewidth}{
\begin{minipage}[t]{\linewidth}
\centering
\includegraphics[width=\linewidth]{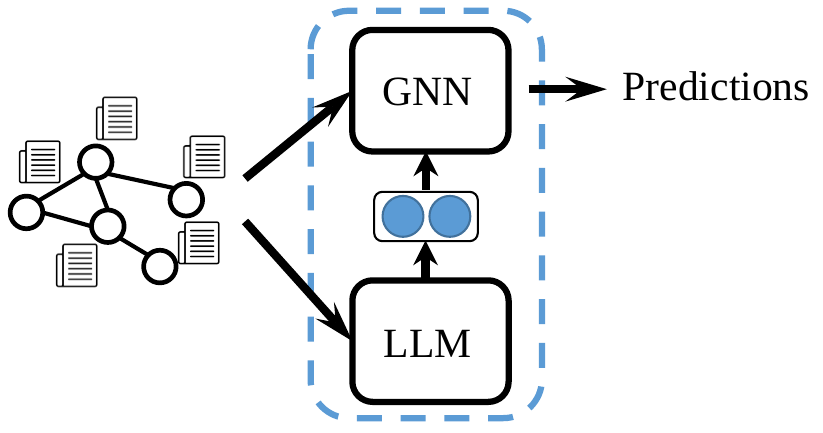}
%\caption{LLM-embedded methods.}
\end{minipage}%
}}%
\hspace{0cm}
\subfigure[Symmetric methods, where the aligned embeddings can be further utilized for downstream tasks.]{
\parbox[][4cm][c]{0.35\linewidth}{
\begin{minipage}[t]{\linewidth}
\centering
\includegraphics[width=\linewidth]{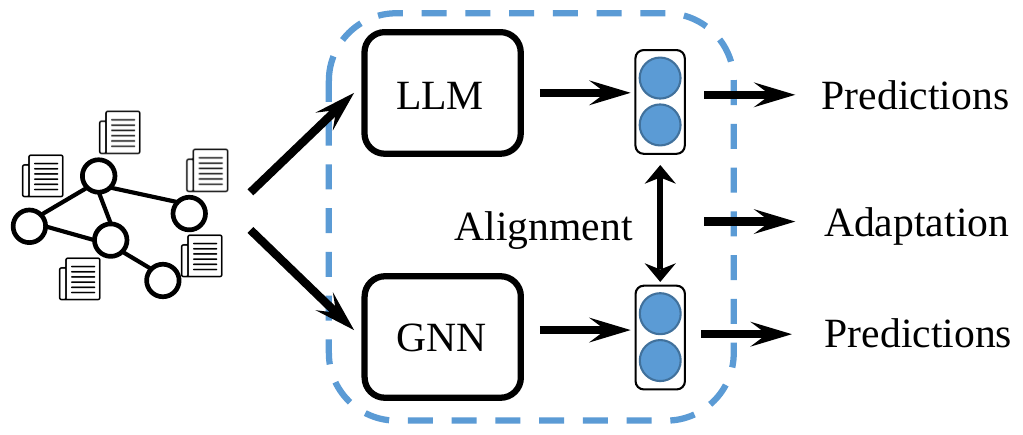}
%\caption{Synergized LLM and GM.}
\end{minipage}%
}}  
\hspace{0cm}
\subfigure[LLM-centric methods, which take an instruction as input and output an answer.]{
\parbox[][4cm][c]{0.28\linewidth}{
\begin{minipage}[t]{\linewidth}
\centering
\includegraphics[width=\linewidth]{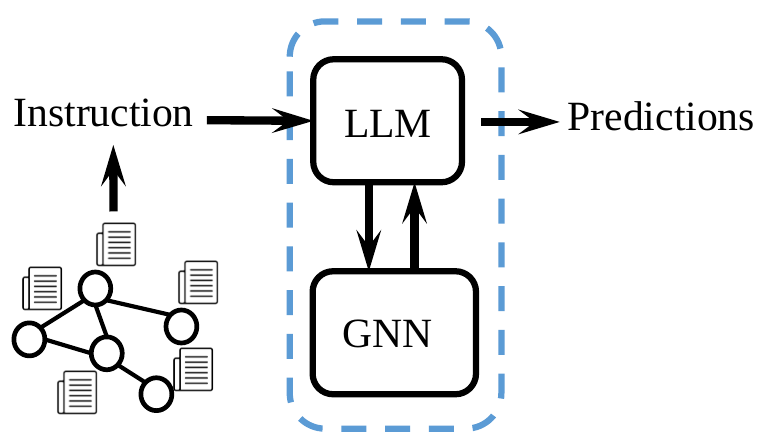}
\end{minipage}}
}
\centering

\caption{An illustration of GNN+LLM-based models.}
\label{fig:LLMGM}
\end{figure*}

GNN-based models lack the ability to process text and thus cannot directly make predictions based on textual data. Additionally, they cannot make predictions based on natural language instructions provided by users. Consequently, exploring the performance of models with a substantial parameter count in graph-related tasks is imperative. On the other hand, LLM-based models for graph learning have their inherent limitations. These limitations include the incapability of LLMs to process precise mathematical calculations and the inability to handle multi-hop logical reasoning, etc. These shortcomings underline the necessity for further research and innovation in this domain. To overcome these limitations and harness the strengths of both language understanding from LLMs and structural analysis from GNNs, integrating LLMs and GNNs can potentially lead to a more comprehensive and powerful model. We summarize and categorize the works mentioned in this section in supplemental material E.%Table ~\ref{tab:llm+GNN}.
%, capable of effectively addressing the intricacies of both text and graph data, thus significantly enhancihttps://www.overleaf.com/project/650546241d9fcbe6c45f36a4ng performance across various tasks

\subsection{Backbone Architectures}
To simultaneously utilize information from both the graph and text and accomplish a variety of tasks, we need to design a framework that effectively integrates LLM and GNN.
Depending on the prediction model, GNN+LLM-based methods can be classified as: 1) GNN-centric Methods, 2) Symmetric Methods, and 3) LLM-centric Methods, as illustrated in Figure~\ref{fig:LLMGM}.

\subsubsection{GNN-centric Methods}
\label{sec:syn-llm-embedded}
Several works aim to utilize LLM to extract node features from raw data and make predictions using GNN. These approaches are denoted as GNN-centric models. For example, GraD ~\cite{GraD} performs a parameter-efficient fine-tuning of an LLM on the textual dataset of a TAG (text-attributed graph). The textual dataset $T$ is annotated with task-specific labels $\mathbf{Y}$, where $G=(V,E,T)$ and $T$ is the set of texts with each element aligned with a node in $V$. Then the downstream task loss for fine-tuning is:

\begin{equation}
	\begin{split}
	&\operatorname{Loss}_{\text{CLS}}=\mathcal{L}_\theta(\phi(\operatorname{LLM}(T)), \mathbf{Y}), \quad \\&\operatorname{Loss}_{\text {LINK }}=\mathcal{L}_\theta\left(\phi\left(\operatorname{LLM}\left(T_{\text {src}}\right), \operatorname{LLM}\left(T_{\text {dst }}\right)\right), \mathbf{Y}\right),
	\end{split}
\end{equation}
where $\phi(\cdot)$ is the classifier for the classification task or similarity function for the link prediction task, $T_{\text{src}}$ and $T_{\text{dst}}$ are the texts of the source node and the target node, respectively, $\operatorname{Loss}_{\text{CLS}}$ and $\operatorname{Loss}_{\text {LINK }}$ are the loss of classification and link prediction task, respectively. Thus we can get the node representations $\mathbf{X}$ with fine-tuned LLM, achieved by removing the head layer. Then we can train GNN with the loss:
%不通顺

\begin{equation}
\resizebox{0.9\hsize}{!}{$
\begin{aligned}
&\operatorname{Loss}_{\text{CLS}}=\mathcal{L}_\theta(\phi(\operatorname{GNN}(\operatorname{LLM}(T))), \mathbf{Y}), \quad \\&\operatorname{Loss}_{\text {LINK}}=\mathcal{L}_\theta\left(\phi\left(\operatorname{GNN}\left(\operatorname{LLM}(T_{\text {src }})\right), \operatorname{GNN}\left(\operatorname{LLM}(T_{\text {dst })}\right)\right), \mathbf{Y}\right),
\end{aligned}$}
\end{equation}
{where $\phi(\cdot)$ is the classifier for the classification task or similarity function for the link prediction task, $\mathbf{Y}$ is the task-specific labels,   $T$ is the set of texts, $T_{\text{src}}$ and $T_{\text{dst}}$ are the texts of the source node and the target node, respectively.}

For LLMs that do not provide direct access to their embeddings such as ChatGPT, TAPE ~\cite{TAPE} engages these LLMs through text interactions. Specifically, TAPE first utilizes an LLM to generate a ranked prediction list and explanation based on the original text, and then an LM is utilized and fine-tuned to transform the original text and additional features of predictions and explanation generated by LLM into node features. Subsequently, downstream GNNs can utilize the features for prediction tasks. 
TAPE extracts graph-agnostic features and cannot capture correlations between graph topology and raw features. To this end, GIANT ~\cite{GIANT} utilizes a graph-structure aware self-supervised learning method to finetune the LM. Consequently, the text representations encompass graph-related information. 
% Concerning the scalability problem of combining LLM and GNN, GraD~\cite{GraD} concurrently optimizes a GNN teacher and a graph-free student over the nodes of the graph using a shared LLM. This encourages the graph-free student to make use of graph information provided by the GNN teacher while allowing the GNN teacher to effectively utilize textual information from unlabeled nodes.
{WTGIA \citep{WTGIA} focuses on text-level Graph Injection Attacks (GIAs), enhancing the interpretability and real-world applicability of graph injection attacks.}
In GALM ~\cite{graphaware}, the focus is on exploring pre-training approaches for models that combine text and graph data, particularly on extensive heterogeneous graphs enriched with rich textual data. OFA \citep{ofa} introduces text-attributed graphs that use natural language to describe nodes and edges, unified by language models into a common embedding space. Heterformer \citep{heterformer} integrates contextualized text encoding and heterogeneous structure encoding within a single model. It incorporates heterogeneous structure information into each Transformer layer as it encodes node texts. Edgeformers \citep{edgeformers}, which are based on graph-enhanced Transformers, perform edge and node representation learning by contextually modeling texts associated with edges. LLMRec \citep{llmrec} improves recommender systems by using three straightforward yet powerful LLM-based graph augmentation techniques, addressing the issues of sparse implicit feedback and low-quality side information commonly found in recommendation systems. 
WalkLM \citep{walklm} conducts attributed random walks on the graph and uses an automated program to generate approximately meaningful textual sequences from these walks. It then fine-tunes a language model (LM) with these textual sequences and extracts embedding vectors from the LM, capturing both attribute semantics and graph structures. TOUCHUP-G \citep{touchup-g} enhances the node features derived from a pre-trained model for downstream graph tasks and introduces. Multiplex graph neural networks initialize node attributes as feature vectors for node representation learning, but they fall short in capturing the full semantics of the nodes' associated texts. METERN \citep{metern} addresses this by using a single text encoder to model the shared knowledge across relations and employing a small number of parameters per relation to generate relation-specific representations. 
% ZeroG \citep{zerog} is a model specifically designed for the efficient training of language models, enabling zero-shot transfer across diverse text-attribute graphs. It addresses inherent challenges such as feature misalignment, mismatched label spaces, and negative transfer. 
Another research~\citep{llmtopological} investigates the use of LLMs to improve graph topological structures, a relatively unexplored area. A label-free pipeline, LLM-GNN \citep{llm-gnn}, uses LLMs for annotation and supplies training signals to GNNs for subsequent prediction.
% \citep{gnngpt} incorporates GPT-4 into GNAS and introduces a novel GPT-4-based Graph Neural Architecture Search method. This method designs a new set of prompts to guide GPT-4 in generating graph neural architectures.

\subsubsection{Symmetric Methods}
\label{sec:syn-uncoupled}
Also, there are some works that align the embeddings of GNN and LLM to make better predictions or utilize the embeddings for other downstream tasks, denoted as symmetric methods.
Most GNN-centric based methods involve two sequential steps: text encoding and graph aggregation. It is important to note that during the generation of text embeddings, there is no exchange of information between nodes. 
%Thus it fails to , which could benefit from mutually reinforcing their underlying semantics. 
To consider the interrelated nature of connected nodes, several works try to utilize GNN and LLM together to get structure-aware text features.
GraphFormer ~\cite{GraphFormer} fuses text embedding and graph aggregation as an iterative workflow. During each iteration, the interconnected nodes will engage in information exchange within the layerwise GNN component, formulated as $\hat{z}^l = \operatorname{GNN}(z^l)$, where $\mathbf{z}^l$ is the output of $l$-th layer of GNN.

As a result, each node will incorporate information from its neighboring nodes. The Transformer component then operates on these enhanced node features, enabling the generation of progressively more informative node representations as $z^{l+1}=\operatorname{TRM}(\operatorname{CONCAT}(\hat{z}^l,h^{l})),$
% \vspace{-2mm}
% \begin{equation}
% \begin{split}
%     &\hat{h^{l}}= \operatorname{Concat}(\hat{z}^l,h^{l}),\\
%     &h^{l+1}=\operatorname{TRM}(\hat{h}^l),\\
%     &z^{l+1}=h^{l+1},
% \end{split}
% \end{equation}
% \vspace{-1mm}
where $\operatorname{TRM}$ is the transformer, and $h^l$ is the output of $l$-th layer of transformer. However, this method suffers from scalability issues because the memory complexity is proportional to the graph size as neighborhood texts are also encoded. GLEM ~\cite{GLEM} employs a variational EM framework to alternatively update the LLMs and GNNs, thus essentially capturing the node label distribution conditioned on the local text attributes. In contrast, GNN uses the text and label information of neighboring nodes to predict labels, thus characterizing global conditional label distribution. By doing so, GLEM efficiently incorporates local textual data and global structural information into its components and can ease the scalability issue.

Other studies employ distinct encoders for graph nodes and texts, training them to align their representations within a shared latent space.
% ConGrat ~\cite{congrat} utilizes a contrastive pre-training framework for jointly learning embeddings of graph nodes and texts. 
G2P2 ~\cite{G2P2} jointly pre-trains a graph-text model utilizing three graph interaction-based contrastive strategies, and then explores prompting for the downstream tasks.
\citep{adaptor} utilizes GNN to model the structural information of nodes, which is then integrated with the corresponding text fragment encoded by a language model. The model subsequently predicts the masked token. 
% SAFER ~\cite{SAFER} trains a text encoder and GNN separately to get the embeddings, and concatenates the embeddings to get the final social context-aware embedding for a logistic regression (LR) classifier. 
ENGINE \citep{engine} integrates large language models and graph neural networks using an adjustable side structure. This approach significantly reduces training complexity while maintaining the capacity of the combined model. To address this, 
PATTON \citep{patton} incorporates two pre-training strategies: network-contextualized masked language modeling and masked node prediction, aiming to capture the inherent relationship between textual attributes and network structure. 
% Grenade \citep{GRENADE} harnesses the combined power of pre-trained language models and graph neural networks for self-supervised representation learning on text-attributed graphs.
{OpenGraph \citep{opengraph} enhances the graph learning paradigm by developing a flexible graph foundation model. This model can understand complex topological patterns in diverse graph data, enabling it to excel in zero-shot graph learning tasks across a range of downstream datasets.} 
{RLMRec \citep{rlmrec} improves the recommendation performance of current recommender systems by utilizing large language models (LLMs) and aligning their semantic space with collaborative relation modeling to achieve better representation learning.} 
Some other works ~\cite{text2mol, MoleculeSTM, CLAMP} also utilize GNN and LLM to learn representations for molecules. These models employ a contrastive learning strategy to effectively pre-train on a dataset containing pairs of molecular graphs and corresponding textual descriptions. By simultaneously learning the chemical structures of molecules and their associated text through this approach, these models can then be applied to various downstream tasks.
Furthermore, MolCA \citep{molca} allows a language model (LM) to comprehend both text-based and graph-based molecular information through the use of a cross-modal projector. GIT-Mol \citep{git-mol} encompasses all three modalities in molecular science—graph, image, and text—supporting tasks such as molecule generation, molecule captioning, molecular image recognition, and molecular property prediction. 
%InstructMol \citep{instructmol} successfully aligns molecular structures with natural language using an instruction-tuning method. 
%This approach employs a two-stage training strategy that skillfully integrates limited domain-specific data with both molecular and textual information.  

% \begin{figure*}
%     \centering
%     \includegraphics[width=0.8\textwidth]{images/GTCL.pdf}
%     \caption{\footnotesize{Illustation of  Graph-Text Contrastive Learning \cite{CLAMP}.}}
%     \label{fig:GTCL}
% \end{figure*}

\subsubsection{LLM-centric Methods}
\label{sec:syn-gm-embedded}
While LLMs have shown impressive performance in various natural language tasks, they struggle with precise mathematical calculations, multi-step logic reasoning, spatial and topological perception, and handling temporal progression. Hence some works utilize GNNs to enhance the performance of LLM, denoted as LLM-centric methods. For example, 
%Graph-ToolFormer ~\cite{zhang2023graph} enhances LLMs by incorporating prompts augmented by ChatGPT, seamlessly integrating API calls from external graph learning tools into graph reasoning statements. This enhancement empowers LLMs to handle graph reasoning tasks more effectively, including fundamental graph property reasoning as well as advanced tasks like reasoning about bibliographic paper topics, molecular graph functions and sequential recommender systems. 
GraphTranslator \citep{graphtranslator} utilizes a Graph Model to efficiently manage predefined tasks and takes advantage of the extended interface of Large Language Models to support a variety of open-ended tasks for the Graph Model.
GraphGPT \citep{graphgpt} integrates large language models with graph structural knowledge through graph instruction tuning, enabling LLMs to understand complex graph structures and improving adaptability across various datasets and tasks.
THLM \citep{thlm} introduces a novel pre-training framework for language models that explicitly incorporates the topological and heterogeneous information found in text-attributed heterogeneous graphs.
% These approaches have not successfully integrated graph information and downstream tasks concurrently. To address this, G-Prompt \citep{g-prompt} combines a graph adapter with task-specific prompts to extract node features effectively. 
GraphPrompter \citep{Graph-Prompter} aligns graph information with LLMs via soft prompts.
% GHGNAS \citep{ghgnas} employs a set of prompts designed to direct GPT-4 in the generation of new heterogeneous graph neural architectures.
% GNNAVI \citep{gnnavi} uses a Graph Neural Network (GNN) layer to accurately direct the aggregation and distribution of information flow while processing prompts, by embedding the desired information flow directly into the GNN.
InstructGraph \citep{instructgraph} enhances LLMs with graph reasoning and generation capabilities through instruction tuning and preference alignment.
RELM \citep{relm} utilizes the chemical knowledge embedded in LMs to support GNNs, thereby improving the accuracy of real-world chemical reaction predictions.
{TEA-GLM \citep{TEA-GLM} pretrains a GNN using contrastive learning to capture structural and semantic graph information, then employs a linear projector to map GNN representations into unified task-specific instructions for LLMs, enabling effective cross-dataset and cross-task generalization without fine-tuning the LLM.
G-Retriever \citep{G-Retriever} introduces G-Retriever, a retrieval-augmented generation (RAG) framework that enables question answering on real-world textual graphs through a conversational interface, mitigating hallucinations and scaling efficiently to large graphs.}

\subsection{Pre-training}

To train the model and enable it to handle both graph and text information, we need to train the model on a large amount of data. LLM and GNN can be pre-trained on textual data and graph data respectively, and the GNN+LLM-based methods can be pre-trained on both data. In this subsection, we category the pre-training strategies as GNN or LLM-based, and alignment-based.

\subsubsection{GNN or LLM-based}
\label{sec:syn-gl}
Other frameworks leverage pre-trained LLMs to obtain text embeddings. The majority of existing models~\cite{GIANT, GraD, GraphFormer, GLEM, text2mol, molca, MoleculeSTM, CLAMP} employ Masked Language Modeling (MLM) during pre-training. Some models, like TAPE and Graph-ToolFormer, opt for Language Modeling (LM) in the pre-training phase.  Additionally, SimTeG integrates Text-Text Contrastive Learning (TTCL), a technique that leverages certain observed text pairs exhibiting more semantic similarity than randomly selected pairs during the pre-training phase as: 
% \begin{small}
    \begin{equation}
    \resizebox{0.9\hsize}{!}{$\operatorname{Loss}_{\operatorname{TTCL}}=\mathbf{E}_{x, y^{+},y^{-}}\left[-\log \frac{\exp \left(k\left(x, y^{+}\right)\right)}{\exp \left(k\left(x, y^{+}\right)\right)+\exp \left(k\left(x, y^{-}\right)\right)}\right],$
    }
\end{equation}
% \end{small}
where $\mathbf{E}$ is the expectation, $k$ is the score function, $y^{+}$ is the positive sample and $y^{-}$ is the negative sample. Additionally, GALM \cite{graphaware} utilizes graph reconstruction for pre-training on extensive graph datasets, and thus can incorporate the graph information into the pre-trained LLMs.

\subsubsection{Alignment-based}
\label{sec:syn-align}
Symmetric methods of LLM and GNN like Text2Mol \cite{text2mol}, MoleculeSTM \cite{MoleculeSTM}, and CLAMP \cite{CLAMP} are pre-trained on large datasets with Graph-Text Contrastive Learning (GTCL), which aligns the embeddings of the graph encoder and the text encoder. The embeddings involve rich information about graph structure and text, thus demonstrating appealing performance on downstream datasets. For a molecule example, CLAMP minimizes the contrastive loss as below:
\begin{equation}
\begin{aligned}
    \text{Loss}_{\text{NCE}}&=-\dfrac{1}{N}\sum_{i=1}^{N}y_i\log(k(\text{LLM}(\mu_i),\text{GNN}(\xi_i)))\\ &+(1-y_i)\log(1-k(\text{LLM}(\mu_i),\text{GNN}(\xi_i))),
\end{aligned}
\end{equation}
where $\mu_i$ is the text representation, $\xi_i$ is the graph representation, and $k$ is a score function to predict the activity of a molecule. The contrastive loss promotes the active molecules on a bioassay to have similar embeddings to the specific bioassay, while ensuring that inactive molecules have dissimilar embeddings to it.
%As shown in Figure~\ref{fig:GTCL}, the contrastive loss promotes the active molecules on a bioassay to have similar embeddings to the embedding of a specific bioassay, while ensuring that inactive molecules have dissimilar embeddings to it.

\subsection{Adaptation}

The adaptation phase plays a pivotal role in optimizing GNN+LLM-based models for efficient graph learning. Apart from some works \cite{text2mol, MoleculeSTM, CLAMP} which test the model's performance on zero-shot tasks such as zero-shot structure-text retrieval and zero-shot text-based molecule editing, models in most cases need adaptation. In this subsection, we categorize these adaptation strategies into two main types: fine-tuning and prompt-tuning. %In the subsequent sections, we will provide a detailed introduction to the specific adaptation strategies employed by these approaches.

\subsubsection{Fine-tuning}
\label{sec:syn-ft}
To adapt to the downstream tasks, some works \cite{ GraphFormer, walklm, heterformer, GIANT, GLEM, edgeformers, graphaware, GraphFormer} utilize vanilla fine-tuning methods for node classification tasks. However, vanilla fine-tuning methods involve adjusting a broad range of model parameters, which can be computationally intensive and resource-demanding. {So other works utilize parameter-efficient fine-tuning methods for downstream tasks, resulting in a more efficient and resource-friendly approach. Specifically, several studies \citep{text2mol, MoleculeSTM, CLAMP} align the embedding space of GNN and LLM utilizing paired molecule graph-text data, while other research 
\citep{GraD, thlm, patton, metern, llmrec, llm-gnn} is tuned on TAGs with classification task. Additionally, some works \citep{graphtranslator, molca, Graph-Prompter} adapt to downstream tasks by generating text captions or descriptions. }

\subsubsection{Prompt-Tuning}
\label{sec:syn-pt}
The prompt-tuning approach is employed in certain studies \cite{G2P2,  adaptor, git-mol,graphgpt, instructgraph}. 
% For Graph-ToolFormer \cite{zhang2023graph}, a substantial dataset of prompts linked to graph reasoning API call statements is annotated and expanded using an LLM, integrating the most suitable calls to external APIs. This dataset is subsequently employed to fine-tune open-source language models like LLaMA, imparting them with the capability to optimally utilize external graph reasoning tools. 
For example, G2P2 \cite{G2P2} leverages prompt-tuning to automatically optimize prompts with limited labeled data for efficient adaptation to downstream tasks.
{Other studies \citep{TAPE, ofa, llmrec, relm} exclusively focus on utilizing Tuning-Free Prompting to generate text. These approaches leverage the inherent capabilities of language models without any additional fine-tuning or parameter adjustments, thereby relying solely on the pre-trained knowledge embedded within the models to produce text outputs.} For example, in TAPE \cite{TAPE}, the initial text features are incorporated into a specialized prompt to interrogate a language model, generating a ranked list of predictions along with explanations. Subsequently, the expanded text features are utilized for finetuning on an LLM. 

\subsection{Discussion}
To summarize, LLMs excel in capturing complex linguistic patterns and semantics from textual data, allowing the GNN+LLM-based models to generate embeddings that involve rich text, structure information, and even external knowledge of LLMs, thus leading to better model performance. Also, when integrated with GNN, LLM's reasoning capabilities over graphs may be enhanced. At the same time, these models can also be regarded as multimodal models to accomplish cross-modal tasks, such as text-graph retrieval tasks. The embeddings can be then utilized for a bunch of downstream tasks. 

% However, training a GNN+LLM-based model often requires large amounts of data (especially paired data) and significant computational resources due to the complexity and scale of both LLMs and GNNs. Effectively integrating LLMs and GNNs in a coherent and effective manner is challenging and requires careful design and experimentation to achieve optimal performance.

{Also, it is challenging to align LLMs and GNNs into a common representational space. To tackle this problem, it is essential to establish a robust standard for measuring the alignment between LLM and GNN representations. This standard should evaluate the degree to which the embeddings from both models capture similar semantic and structural information. Additionally, it is crucial to design effective methods for achieving this alignment. By doing so, we can ensure that the combined model leverages the strengths of both LLMs and GNNs, ultimately enhancing performance on various downstream tasks.}
{Moreover, existing work has begun to extend the GNN+LLM approach to heterogeneous graphs and hypergraphs. For heterogeneous graphs, HiGPT \citep{tang2024higpt} introduces an in-context heterogeneous graph tokenizer and a heterogeneity-aware instruction-tuning framework to address distribution shifts, enhancing generalization and performance across various heterogeneous graph learning scenarios, and GHGRL~\citep{GHGRL} employs LLM to automatically summarize and classify different data formats and types of heterogeneous graph data. For hypergraph, HyperBERT \citep{hyperbert2024} augments a pre-trained BERT model with specialized hypergraph-aware layers for the task of node classification.}
% zmm add：基于clip的，这种都是很明显的contrastive learning的
% 详见：https://arxiv.org/pdf/2303.04226.pdf 里的Fig.13

\section{Challenges and Future Directions}
\label{sec:future}
Although the previous sections have discussed the concepts and a lot of related works towards graph foundation models, there are still many avenues for future exploration in this research area. This section will delve into these issues.
% Impact 
% (1) 确定有知识 Graph Domain*
% 1.transferablility (node level=迁移性不强， graph level=生物分子) 
% 2.有待挖掘
% （2）如何构建高质量数据：
% https://www.overleaf.com/project/650546241d9fcbe6c45f36a41.ChatGPT, data centric, scale, quality
% 2.--->pipeline
%\subsubsection{When graph foundation model meets data-centric graph learning: Substitutes or Complements?}
% LLM出现后，数据挖掘(Data Mining)可以做哪些新的研究？https://www.zhihu.com/question/607312619/answer/3229062973
% \subsubsection{How to improve graph foundation model via data-centric machine learning?}

% why: ChatGPT效果好的原因之一是数据量和数据质量提升，离不开data-centric，导致emergence。
% how: 现在有一些图data-centric的工作，但是还没和图基础模型结合。language有一套语义，但在graph上没有，所以有Challenge。

%Data-centric Graph Learning emphasizes the importance of the data itself in graph-based machine learning. It is concerned with data preprocessing techniques, data cleaning, handling missing data, addressing noise, dealing with dynamic graphs, and ensuring that the graph data used for training and evaluation is of high quality and representative of the real-world phenomena being modeled. 

% 大的问题：scope, variety, noisy?
\subsection{Challenges about Data and Evaluation}

\subsubsection{Data Quantity and Quality}
The improvements in data quantity and data quality are the key factors contributing to the effectiveness of foundation models~\cite{bommasani2021opportunities}. At present, there is still a limited amount of open-source large-scale graph data~\cite{hu2020open,khatua2023igb}, and each dataset is primarily concentrated in a single domain. This poses a challenge to learn graph foundation models for diverse data domains. Hence, it is necessary to collect and organize a unified, massive dataset that covers graph data and related corpora across different domains. {Note that some works have constructed cross-domain datasets~\cite{li2024teg, li2024can}, which aid in developing cross-domain graph foundation models.} Additionally, if the graph data are noisy, incomplete, or not properly curated, it will negatively affect the performance of graph foundation models. 
To enhance the data quality of GFMs, efforts have been made to propose augmentation strategies from various perspectives, including 
graph structure learning, feature competion and label mixing, etc. 
%graph augmentation~\cite{GraphCL}, feature augmentation~\cite{yang2021graph}, label augmentation~\cite{sun2020multi}, etc. 
%However, these approaches typically apply to homogeneous graphs. When the graph types extend to complex forms like heterogeneous graphs or dynamic graphs, improving data quality for graph foundation models becomes a significant challenge. 
However, since existing data augmentation techniques are typically tailored for individual GNN-based models, there is a need for further exploration on how to effectively augment graph data for LLM-based or GNN+LLM-based models. 

% 如何收集、组织一个统一的超大图数据集，覆盖同域甚至跨域的图数据和相关语料，也是一个重要的方向
\subsubsection{Evaluation}
%~\cite{chang2023survey}
With the help of natural language instructions and powerful generation capabilities, LLMs can support a variety of open-ended tasks~\cite{LLaMA}. This presents new opportunities for graph foundation models based on LLM. However, due to the lack of labels in open-ended tasks, evaluating the performance of GFMs in such tasks is a challenge. When using LLM as a language foundation model, the evaluation of its performance on open-ended tasks has evolved from human evaluation~\cite{ouyang2022training} to meta-evaluation~\cite{zeng2023evaluating}. The question of whether existing LLM evaluation methods~\cite{ouyang2022training,zeng2023evaluating} can be applied to GFMs remains to be explored. 
%In particular, the diversity in data domains and task formats for graph data poses greater challenges for the evaluation of graph foundation models. For GNN-based or LLM-based graph foundation models, the question of whether existing GNN evaluation methods~\cite{shchur2018pitfalls,yin2023dgi} and LLM evaluation methods~\cite{ouyang2022training,zeng2023evaluating} can be applied remains to be explored. For GNN+LLM-based graph foundation models, devising an appropriate evaluation approach poses a more significant issue for investigation. 
Beyond evaluating the performance of GFMs, it is also worth evaluating their robustness, trustworthiness, or holistic performance, similar to the current practices for language foundation models~\cite{wang2021adversarial, wang2023decodingtrust, bommasani2023holistic}.

\subsection{Challenges about Models}
\subsubsection{Model Architectures}
As mentioned above, the designs of backbone architectures and learning paradigms are crucial for the implementation of GFMs. Although this article has outlined some potential solutions, it does not rule out the possibility of better ones. For example, regarding the backbone architecture, recent works have proposed model architectures that go beyond the Transformer, offering improved performance~\cite{dao2024transformers} or interpretability~\cite{yu2023white}. However, it is still unknown whether these backbone architectures can be used for dealing with graph data. Additionally, when utilizing GNN+LLM-based models, it is worth exploring how to more effectively align the outputs of both models. Furthermore, there is limited research regarding the emergent abilities or neural scaling law~\cite{wang2024exploring, maoposition} of GNN-based~\cite{PRODIGY} or LLM-based~\cite{NLGraph} graph foundation models. It is yet unclear whether GNN+LLM-based models may have greater potential for emergence. {Furthermore, considering the diverse types of graphs (such as heterogeneous graphs~\cite{tang2024higpt}, temporal graphs~\cite{zhang2023llm4dyg}, and hypergraphs~\cite{hyperbert2024}), designing a GFM capable of handling multiple types of graphs is a valuable direction for future research. A potential solution is to use a mixture of experts (MoE) model~\cite{wang2023graph}, where each expert handles one type of graph.} Finally, given that current multimodal foundation models~\cite{fei2022towards} primarily handle text, images, audio, and other modalities, it is an interesting research direction to explore whether GNNs can be employed to further expand the diversity of modalities covered by multimodal foundation models or enhance the capabilities of foundation models for multimodal learning~\cite{yuan2023power}. 

% 涌现：增加参数量或者和LLM结合
% 图作为一类模态数据，能否提升多模态大模型？目前多模态大模型大多没考虑图数据，图数据能否有用，如何融入。

\subsubsection{Model Training}
In order to achieve homogeneity and make effective use of pre-training data, it is crucial to design appropriate pre-training tasks in pre-training. {Unlike many language foundation models, which often use LM~\cite{ouyang2022training} or MLM~\cite{devlin2018bert} as pre-training tasks, there are now various forms of pre-training tasks tailored to different GFM model architectures. 
Whether each type of pre-training task has its own applicable scope and whether there will be a unified pre-training task are worth further exploration.} {Additionally, enabling graph foundation models to support cross-domain data is a vital concern. Some works use data from different domains as model input for pre-training~\cite{davies2023its, zhao2024all}, or enable adaptation to data from different domains through methods such as LLM-based embedding~\cite{ofa}, condition generation~\cite{zhu2024graphcontrol} and zero-shot transfer~\cite{tang2024higpt}.} Finally, apart from fine-tuning and prompting that are introduced in this article, there are other potential training techniques that can be applied to improve efficiency or update knowledge, such as knowledge distillation~\cite{jiang2023lion}, reinforcement learning from human feedback (RLHF)~\cite{ouyang2022training} and model editing~\cite{yao2023editing}. Whether the above-mentioned techniques can be applied to graph foundation models will be a focal point of future research.

% 同质性  由于图数据和任务的差异性使得同质性还没有显现。图里面的通用任务是什么？任务的本质共性是什么？跨域任务的内在一致性是什么？怎么实现同质性。 人类的行为一致性是否也是一种方案。涌现和同质性可以放在一起。

\subsection{Challenges about Applications}

\subsubsection{Killer Applications}
In comparison to the outstanding performance of language foundation models in tasks like text translation~\cite{hendy2023good} and text generation~\cite{zhang2023controllable}, whether GFMs can similarly catalyze groundbreaking applications in graph tasks is not yet clear. 
For scenarios that are well-suited for the application of GNNs, such as e-commerce~\cite{zhang2020agl} and finance~\cite{wang2021review}, potential research directions include leveraging graph-based models integrated with LLMs to better support open-ended tasks~\cite{graphtranslator}, or enhancing the reasoning capabilities of LLMs through graph learning techniques~\cite{yu2023thought}. Furthermore, GFMs have the potential to make breakthroughs in some emerging fields. For example, drug development is a time-consuming and costly process~\cite{wouters2020estimated}, and language foundation models have already been successfully used for related tasks like target identification and side effect prediction~\cite{bommasani2021opportunities}. Given the 3D geometric structure of proteins~\cite{liu2021pre}, GFMs hold the promise of enhancing the drug discovery pipeline by leveraging their ability to model graph structure information~\cite{xia2022systematic}, potentially speeding up the process further. Additionally, urban computing may also represent a crucial application scenario for GFMs. It is worth noting that traditional traffic prediction techniques have been primarily focused on addressing individual tasks such as travel demand prediction~\cite{zhuang2022uncertainty} and traffic flow prediction~\cite{wang2020traffic}, lacking a comprehensive understanding of the entire transportation system. Given that the transportation system can be viewed as a spatio-temporal graph, graph foundation models hold the potential to capture the participation behavior of actors in the transportation system~\cite{wang2023building}, thereby offering a unified approach to addressing various issues in urban computing.%, lin2023comprehensive, lin2022effectively

%, such as drug discovery~\cite{xia2022systematic} and urban computing~\cite{wang2023building},
% instruction following
% \textbf{Emergent abilities}

\subsubsection{Trustworthiness}

Despite the strong performance of LLM-based foundation models, their black-box nature~\cite{sun2022black} introduces a host of safety concerns, such as hallucination and privacy leaks. The hallucination refers to the output appearing plausible but deviating from user input, context, or facts~\cite{zhang2023siren}. Existing research suggests that this phenomenon is associated with multiple factors, such as the model's overconfidence in its own behavior~\cite{ren2023investigating} and the misunderstanding of false correlations~\cite{li2022pre}. Similarly, recent work has pointed out that pre-trained GNNs also pose certain trustworthy risks about fairness~\cite{zhang2024endowing} and robustness against attacks~\cite{zhang2022inference, zhang2024can}. Given the unique nature of graph data, we may require certain techniques to prevent or mitigate security risks on GFMs, such as confidence calibration~\cite{wang2021confident} or counterfactual reasoning~\cite{tan2022learning}. Additionally, given that existing research has indicated privacy risks in both GNN~\cite{wu2022federated, yan2024federated} and LLM~\cite{staab2023beyond}, enhancing the privacy of GFMs is also a critical concern. Some potential solutions include federated learning~\cite{chen2023privacy}, RLHF~\cite{ouyang2022training} and red teaming~\cite{shi2024red}, but whether these methods can be applied to GFMs is still unknown. {Finally, graph data in real-world applications frequently encounter challenges such as noise~\cite{fatemi2021slaps}, class imbalance~\cite{zeng2023imgcl}, data incompleteness~\cite{tu2024revisiting}, and multi-modal features~\cite{lian2023gcnet}. Developing methods to utilize these graph data for constructing GFMs, or adapting existing GFMs to accommodate these characteristics, will be a critical area of focus for future research.}

\section{Conclusions}
\label{sec:conclusions}

The development of foundation models and graph machine learning has spurred the emergence of a new research direction, with the aim to train on broad graph data and apply it to a wide range of downstream graph tasks.
In this article, we propose the concept of graph foundation models (GFMs) for the first time, and provide an introduction to relevant concepts and representative methods. We summarize existing works towards GFMs into three main categories based on their reliance on graph neural networks (GNNs) and large language models (LLMs): GNN-based models, LLM-based models, and GNN+LLM-based models. For each category of methods, we introduce their backbone architectures, pre-training, and adaptation strategies separately. After providing a comprehensive overview of the current landscape of graph foundation models, this article also points out the future directions for this evolving field.

\section*{Acknowledgements}

This work was supported by the National Natural Science Foundation of China (No.U20B2045, 62192784, 62236003), Young Elite Scientists Sponsorship Program (No.2023QNRC001) by CAST, NSF under grants III-2106758 and POSE-2346158.

% \clearpage

% \bibliographystyle{ieeetr.bst}
% \bibliography{main}
% \bibliographystyle{IEEEtran}
\small
\bibliographystyle{IEEEtran}
\bibliography{main}
%\clearpage

% \begin{IEEEbiography}[{\includegraphics[width=1in,height=1.25in,clip,keepaspectratio]{./images/jiaweiliu.jpg}}]{Jiawei Liu} received the B.S. degree in computer science and technology from Beijing University of Posts and Telecommunications, Beijing, China, in 2020. He is currently pursuing the Ph.D. degree in computer science and technology from Beijing University of Posts and Telecommunications, Beijing, China. His research interests include graph data mining and machine learning.
% %\begin{IEEEbiographynophoto}{Jun Liu}
% \end{IEEEbiography}
% \vspace{-1 cm} 

\begin{IEEEbiographynophoto}{Jiawei Liu} received the B.S. degree in computer science and technology from Beijing University of Posts and Telecommunications, Beijing, China, in 2020. He is currently pursuing the Ph.D. degree in computer science and technology from Beijing University of Posts and Telecommunications, Beijing, China. His research interests include graph data mining and machine learning.
\end{IEEEbiographynophoto}
\vspace{-1 cm} 

% \begin{IEEEbiography}[{\includegraphics[width=1in,height=1.25in,clip,keepaspectratio]{./images/chengyang.jpg
% }}]{Cheng Yang} is an Associate Professor of Computer Science at Beijing University of Posts and Telecommunications (BUPT). He received his Bachelor and Ph.D. degrees from Tsinghua University in 2014 and 2019, respectively. Cheng's research interests include data mining, natural language processing and social computing. He has published 60+ papers in top journals and conferences, such as NeurIPS, ICLR, KDD and ACL. His work has got more than 10,000 citations as shown by Google Scholar. Cheng is named by Baidu as one of the Top 100 Chinese Young Scholars in Artificial Intelligence.
% %\begin{IEEEbiographynophoto}{Jun Liu}
% \end{IEEEbiography}
% \vspace{-1 cm} 

% \begin{IEEEbiographynophoto}{Cheng Yang} is an Associate Professor of Computer Science at Beijing University of Posts and Telecommunications (BUPT). He received his Bachelor and Ph.D. degrees from Tsinghua University in 2014 and 2019, respectively. Cheng's research interests include data mining, natural language processing and social computing. He has published 60+ papers in top journals and conferences, such as NeurIPS, ICLR, KDD and ACL. His work has got more than 10,000 citations as shown by Google Scholar. Cheng is named by Baidu as one of the Top 100 Chinese Young Scholars in Artificial Intelligence.
\begin{IEEEbiographynophoto}{Cheng Yang} is an Associate Professor of Computer Science at Beijing University of Posts and Telecommunications (BUPT). He received his Bachelor and Ph.D. degrees from Tsinghua University in 2014 and 2019, respectively. Cheng's research interests include data mining, natural language processing and social computing. He has published 60+ papers in top journals and conferences, such as NeurIPS, ICLR, KDD and ACL.
\end{IEEEbiographynophoto}
\vspace{-1 cm} 

% \begin{IEEEbiography}[{\includegraphics[width=1in,height=1.25in,clip,keepaspectratio]{./images/zhiyuanlu.jpg
% }}]{Zhiyuan Lu} received the BS degree in communication engineering from Beijing University of Posts and Telecommunications, Beijing, China, in 2022. He is currently pursuing the PhD degree in computer science and technology from Beijing University of Posts and Telecommunications, Beijing, China. His current research interests are in graph neural networks, data mining and machine learning.
% %\begin{IEEEbiographynophoto}{Jun Liu}
% \end{IEEEbiography}
% \vspace{-1 cm} 

\begin{IEEEbiographynophoto}{Zhiyuan Lu} received the BS degree in communication engineering from Beijing University of Posts and Telecommunications, Beijing, China, in 2022. He is currently pursuing the PhD degree in computer science and technology from Beijing University of Posts and Telecommunications, Beijing, China. His current research interests are in graph neural networks, data mining and machine learning.
\end{IEEEbiographynophoto}
\vspace{-1 cm} 

% \begin{IEEEbiography}[{\includegraphics[width=1in,height=1.25in,clip,keepaspectratio]{./images/junzechen.jpg
% }}]{Junze Chen} is a master's student at the School of Computer Science, Beijing University of Posts and Telecommunications (BUPT). He obtained his Bachelor's degree in Computer Science and Technology from BUPT in 2022. His primary research interests are in data mining, graph neural networks, and natural language processing.
% %\begin{IEEEbiographynophoto}{Jun Liu}
% \end{IEEEbiography}
% \vspace{-1 cm} 

\begin{IEEEbiographynophoto}{Junze Chen} is a master's student at the School of Computer Science, Beijing University of Posts and Telecommunications (BUPT). He obtained his Bachelor's degree in Computer Science and Technology from BUPT in 2022. His primary research interests are in data mining, graph neural networks, and natural language processing.
\end{IEEEbiographynophoto}
\vspace{-1 cm} 

% \begin{IEEEbiography}[{\includegraphics[width=1in,height=1.25in,clip,keepaspectratio]{./images/yiboli.jpg
% }}]{Yibo Li} received the B.S. degree from the Beijing University of Posts and Telecommunications, China, in 2022. She is currently working toward the master’s degree with the Beijing University of Posts and Communications, China. Her current research interests are in graph neural networks and large language models.
% %\begin{IEEEbiographynophoto}{Jun Liu}
% \end{IEEEbiography}
% \vspace{-1 cm}

\begin{IEEEbiographynophoto}{Yibo Li} received the B.S. degree from the Beijing University of Posts and Telecommunications, China, in 2022. She is currently working toward the master’s degree with the Beijing University of Posts and Communications, China. Her current research interests are in graph neural networks and large language models.
\end{IEEEbiographynophoto}
\vspace{-1 cm} 

% \begin{IEEEbiography}[{\includegraphics[width=1in,height=1.25in,clip,keepaspectratio]{./images/mengmeizhang.jpg
% }}]{Mengmei Zhang} received her Ph.D. degree in computer science and technology from Beijing University of Posts and Telecommunications in 2023. She is currently a senior researcher at China Telecom Bestpay. Her research interests include graph mining, large language models, and risk control.
% %\begin{IEEEbiographynophoto}{Jun Liu}
% \end{IEEEbiography}
% \vspace{-1 cm}

\begin{IEEEbiographynophoto}{Mengmei Zhang} received her Ph.D. degree in computer science and technology from Beijing University of Posts and Telecommunications in 2023. She is currently a senior researcher at China Telecom Bestpay. Her research interests include graph mining, large language models, and risk control.
\end{IEEEbiographynophoto}
\vspace{-1 cm} 

% \begin{IEEEbiography}[{\includegraphics[width=1in,height=1.25in,clip,keepaspectratio]{./images/tingbai.jpg
% }}]{Ting Bai} received her Ph.D. degree from Renmin University of China in 2019. She currently is an associate professor in the School of Computer Science, Beijing University of Posts and Telecommunications. Her major research interests are in recommender systems and Human behavior analysis. She has published several papers on SIGIR, WWW, KDD, CIKM, WSDM, TKDE, and so on.
% \end{IEEEbiography}
% \vspace{-1 cm}

\begin{IEEEbiographynophoto}{Ting Bai} received her Ph.D. degree from Renmin University of China in 2019. She currently is an associate professor in the School of Computer Science, Beijing University of Posts and Telecommunications. Her major research interests are in recommender systems and Human behavior analysis. She has published several papers on SIGIR, WWW, KDD, CIKM, WSDM, TKDE, and so on.
\end{IEEEbiographynophoto}
\vspace{-1 cm} 

% \begin{IEEEbiography}[{\includegraphics[width=1in,height=1.25in,clip,keepaspectratio]{./images/yuanfang.jpg
% }}]{Yuan Fang} received the bachelor’s degree in computer science from the National University of Singapore in 2009 and the Ph.D. degree in computer science from the University of Illinois at Urbana-Champaign in 2014. He is currently an assistant professor with the School of Computing and Information Systems, Singapore Management University. His current research focuses on graph-based data mining and machine learning, and their applications.
% \end{IEEEbiography}
% \vspace{-1 cm}

% \begin{IEEEbiographynophoto}{Yuan Fang} received the bachelor’s degree in computer science from the National University of Singapore in 2009 and the Ph.D. degree in computer science from the University of Illinois at Urbana-Champaign in 2014. He is currently an assistant professor with the School of Computing and Information Systems, Singapore Management University. His current research focuses on graph-based data mining and machine learning, and their applications.

\begin{IEEEbiographynophoto}{Yuan Fang} received the Ph.D. degree in computer science from the University of Illinois at Urbana-Champaign in 2014. He is currently an assistant professor with the School of Computing and Information Systems, Singapore Management University. His current research focuses on graph data mining, machine learning and their applications.
\end{IEEEbiographynophoto}
\vspace{-1 cm} 

% \begin{IEEEbiography}[{\includegraphics[width=1in,height=1.25in,clip,keepaspectratio]{./images/lichaosun.jpg
% }}]{Lichao Sun} is an Assistant Professor in Computer Science and Engineering at Lehigh University. He obtained M.S. and B.S. from University of Nebraska Lincoln, and received his Ph.D. degree in Computer Science at University of Illinois Chicago in 2020. His research interests are Trustworthy AI and Medical AI in various applications. He was the recipients of 2024 Microsoft Accelerate Foundation Models Research Award, 2024 OpenAI Researcher Award and NSF CRII Award. He have published more than 90 papers in top-tier journals and conferences, such as Nature Medicine, NeurIPS, ICML, ICLR, AAAI and IJCAI.
% \end{IEEEbiography}
% \vspace{-1 cm}

% \begin{IEEEbiographynophoto}{Lichao Sun} is an Assistant Professor in Computer Science and Engineering at Lehigh University. He obtained M.S. and B.S. from University of Nebraska Lincoln, and received his Ph.D. degree in Computer Science at University of Illinois Chicago in 2020. His research interests are Trustworthy AI and Medical AI in various applications. He was the recipients of 2024 Microsoft Accelerate Foundation Models Research Award, 2024 OpenAI Researcher Award and NSF CRII Award. He have published more than 90 papers in top-tier journals and conferences, such as Nature Medicine, NeurIPS, ICML, ICLR, AAAI and IJCAI.

\begin{IEEEbiographynophoto}{Lichao Sun} is an Assistant Professor in Computer Science and Engineering at Lehigh University. He obtained his Ph.D. degree in Computer Science at University of Illinois Chicago in 2020. His research interests are Trustworthy AI and Medical AI in various applications. He have published more than 90 papers in top-tier journals and conferences, such as Nature Medicine, NeurIPS, ICML, ICLR, AAAI and IJCAI.
\end{IEEEbiographynophoto}
\vspace{-1 cm} 

% \begin{IEEEbiography}[{\includegraphics[width=1in,height=1.25in,clip,keepaspectratio]{./images/PhilipS Yu.jpg
% }}]{Philip S. Yu} is a Distinguished Professor at the University of Illinois Chicago, holding the Wexler Chair in information and Technology. Previously, he managed the Software Tools and Techniques department at IBM Thomas J. Watson Research Center. He is a Fellow of the ACM and IEEE. He has published over 2,000 papers, has more than 19,9800 citations with an H-index of 198, and holds or has applied for over 300 US patents. His main research interests include big data, data mining, privacy preserving publishing and mining, data streams, database systems, Internet applications and technologies.
% \end{IEEEbiography}
% \vspace{-1 cm}

% \begin{IEEEbiographynophoto}{Philip S. Yu} is a Distinguished Professor at the University of Illinois Chicago, holding the Wexler Chair in information and Technology. Previously, he managed the Software Tools and Techniques department at IBM Thomas J. Watson Research Center. He is a Fellow of the ACM and IEEE. He has published over 2,000 papers, has more than 19,9800 citations with an H-index of 198, and holds or has applied for over 300 US patents. His main research interests include big data, data mining, privacy preserving publishing and mining, data streams, database systems, Internet applications and technologies.

\begin{IEEEbiographynophoto}{Philip S. Yu} is a Distinguished Professor at the University of Illinois Chicago, holding the Wexler Chair in information and Technology. He is a Fellow of the ACM and IEEE. He has published over 2,000 papers and holds or has applied for over 300 US patents. His main research interests include big data, data mining, privacy preserving publishing and mining, data streams, database systems, Internet applications and technologies.
\end{IEEEbiographynophoto}

\vspace{-1 cm} 

% \begin{IEEEbiography}[{\includegraphics[width=1in,height=1.25in,clip,keepaspectratio]{./images/chuanshi.jpg
% }}]{Chuan Shi} received the B.S. degree from the Jilin University in 2001, the M.S. degree from the Wuhan University in 2004, and Ph.D. degree from the lCT of Chinese Academic of Sciences in 2007. He joined the Beijing University of Postsand Telecommunications as a lecturer in 2007 and is a professor and deputy director of Beiiing Key Lab of Intelligent Telecommunications Software and Multimedia at present. His research interests are in data mining and machine learning. He has published more than 100 papers in refereed journals and conferences, such as TPAMI, TKDE, KDD, WWW, NeurlPS, ICML and ICLR.
% \end{IEEEbiography}
% \vspace{-1.7 cm}

% \begin{IEEEbiographynophoto}{Chuan Shi} received the B.S. degree from the Jilin University in 2001, the M.S. degree from the Wuhan University in 2004, and Ph.D. degree from the lCT of Chinese Academic of Sciences in 2007. He joined the Beijing University of Posts and Telecommunications as a lecturer in 2007 and is a professor and deputy director of Beiiing Key Lab of Intelligent Telecommunications Software and Multimedia at present. His research interests are in data mining and machine learning. He has published more than 100 papers in refereed journals and conferences, such as TPAMI, TKDE, KDD, WWW, NeurlPS, ICML and ICLR.

\begin{IEEEbiographynophoto}{Chuan Shi} received the Ph.D. degree from the lCT of Chinese Academic of Sciences in 2007. He joined the Beijing University of Posts and Telecommunications in 2007 and is a professor and deputy director of Beijing Key Lab of Intelligent Telecommunications Software and Multimedia at present. His research interests are in data mining and machine learning. He has published more than 100 papers in refereed journals and conferences, such as TPAMI, TKDE, KDD, WWW and NeurIPS.
\end{IEEEbiographynophoto}

\clearpage
% \vspace{3mm}
\noindent\textbf{\large Supplemental Materials:}
\setcounter{equation}{0}
\setcounter{figure}{0}
\setcounter{table}{0}
\setcounter{page}{1}
\makeatletter
\normalsize
\begin{table*}[ht]
\centering
\caption{\footnotesize{Details of approaches involved as GNN-based models.}}
\resizebox{0.9\linewidth}{!}{
\begin{tabular}{@{}llll@{}}
\toprule
Model      & Backbone Architecture & Pre-training & Adaptation \\ \midrule
All In One~\cite{allinone}& GCN, GAT, Graph Transformer        & Same-Scale CL        & Prompt Tuning   \\
PRODIGY~\cite{PRODIGY} & GCN, GAT &Graph Reconstruction, Supervised &Prompt Tuning \\
DGI~\cite{DGI} & GCN      & Cross-Scale CL        & Parameter-Efficient FT  \\
GRACE~\cite{GRACE} & GCN     & Same-Scale CL        & Vanilla FT  \\
FUG~\cite{zhao2024fug} & GCN     & Same-Scale CL        & Vanilla FT  \\
VGAE~\cite{VGAE} & GCN & Graph Reconstruction, Property Prediction & Vanilla FT\\
MA-GCL~\cite{gong2023ma} & GCN & Same-Scale CL & Vanilla FT\\
MultiGPrompt~\cite{yu2024multigprompt} & GCN & Cross-Scale CL, Graph Reconstruction & Prompt Tuning \\
IGAP~\cite{yan2024inductive} & GCN, GAT, GraphSAGE & Same-Scale CL, Cross-Scale CL, Graph Reconstruction & Prompt Tuning \\
HGPROMPT~\cite{yu2024hgprompt} & GCN, GAT, SimpleHGN & Graph Reconstruction & Prompt Tuning\\
GraphMAE~\cite{GraphMAE} & GAT     & Graph Reconstruction         & Parameter-Efficient FT  \\
GraphMAE2~\cite{GraphMAE2} & GAT     & Graph Reconstruction        & Parameter-Efficient FT  \\
GPPT~\cite{sun2022gppt} & GraphSAGE       & Graph Reconstruction, Cross-Scale CL        & Prompt Tuning   \\
VPGNN~\cite{wen2023voucher} & GraphSAGE   & Cross-Scale CL & Prompt Tuning \\
GPT-GNN~\cite{GPT-GNN} & HGT     &  Graph Reconstruction        & Vanilla FT  \\
PT-HGNN~\cite{jiang2021pre} & HGT & Same-Scale CL & Vanilla FT \\
CPT-HG~\cite{jiang2021contrastive} & HGT & Same-Scale CL & Vanilla FT\\
GraphPrompt~\cite{liu2023graphprompt} & GIN       & Graph Reconstruction        & Prompt Tuning  \\
IHP~\cite{yang2024instruction} & PHC       & Graph Reconstruction        & Prompt Tuning  \\
GraphPrompt+~\cite{yu2023generalized} & GIN & Graph Reconstruction, Cross-Scale CL, Same-Scale CL & Prompt Tuning \\
ProNoG+~\cite{pronog} & FAGCN & Same-Scale CL & Prompt Tuning \\
GCC~\cite{qiu2020gcc} & GIN       & Same-Scale CL         & Vanilla FT  \\
GraphCL~\cite{GraphCL} & GIN       & Same-Scale CL         & Parameter-Efficient FT  \\
AdapterGNN~\cite{adaptergnn} & GIN     & Cross-Scale CL, Graph Reconstruction, Same-Scale CL         & Parameter-Efficient FT  \\
PhyGCN~\cite{deng2024pretrained} & HyperGCN & Graph Reconstruction & Parameter-Efficient FT  \\
GPT-ST~\cite{li2023generative} & GPT-ST & Graph Reconstruction & Parameter-Efficient FT  \\
GraphST~\cite{zhang2023spatial} & GraphST & Same-Scale CL &  Parameter-Efficient FT  \\
AAGOD~\cite{guo2023data} & GIN & Same-Scale CL, Supervised& Prompt Tuning\\
GPF~\cite{fang2022universal} & GIN &Cross-Scale CL, Graph Reconstruction &Prompt Tuning \\
%SGL-PT~\cite{zhu2023sgl} & GIN & Same-Scale CL, Graph Reconstruction & Prompt Tuning\\
GCOPE~\cite{zhao2024all} & FAGCN & Same-Scale CL & Prompt Tuning\\
FOTOM~\cite{davies2023its} & GIN & Same-Scale CL & Parameter-Efficient FT\\
TPP~\cite{niu2024replay} & SGC & Same-Scale CL & Prompt Tuning\\
GraphControl~\cite{zhu2024graphcontrol} & GIN & Same-Scale CL & Parameter-Efficient FT\\
G-TUNING~\cite{sun2024fine} & GIN & Same-Scale CL, Graph Reconstruction & Customized FT\\
Graph-BERT~\cite{zhang2020graph} & Graph Transformer     & Graph Reconstruction, Supervised         & Vanilla FT \\
GROVER~\cite{rong2020self} & Graph Transformer     & Property Prediction         & Vanilla FT \\
G-Adapter~\cite{G-adapter} & Graph Transformer      &  Supervised, Graph Reconstruction, Property Prediction       & Parameter-Efficient FT  \\
\bottomrule
\end{tabular}}

\label{tab:gm}
\end{table*}

\subsection*{A. Impact of Graph Data on GFMs}
\label{sec:impact-graph-data}

% \usepackage{booktabs}
\begin{table*}[ht]
\centering
\caption{\footnotesize{Details of approaches involved as LLM-based models.}}
\resizebox{0.98\linewidth}{!}{
\begin{tabular}{@{}llllll@{}}
\toprule
Model       & \multicolumn{3}{l}{Backbone Architecture}                                                                & Pre-training & Adaptation                  \\ \midrule
GIMLET \cite{zhao2023gimlet} & Graph-to-token & + & Transformer & - & - \\
InstructGLM\cite{InstructGLM} & Graph-to-token &+  & Flan-T5/LLaMA                                                                       & MLM,LM       & Manual Prompt Tuning     \\
NLGraph\cite{NLGraph}     & Graph-to-text &+   & GPTs                                                                                & LM           & Manual Prompt Tuning \\
TextForGraph \cite{wenkel2023pretrained}  & Graph-to-text &+ &GPTs  &LM  & Manual Prompt Tuning \\
When\&Why \cite{huang2023can}  & Graph-to-text &+ & GPTs  & LM  & Maunal Prompt Tuning \\
GraphWiz \cite{chen2024graphwiz} & Graph-to-text &+ & LLaMA, Mistral  & LM & Maunal Prompt Tuning \\
CGForLLM \cite{antonucci2023zero} & Graph-to-text &+ & GPT4 & LM & Maunal Prompt Tuning \\
LLM4DYG \cite{zhang2023llm4dyg} & Graph-to-text &+ & LLaMA, Vicuna, GPT-3.5 &LM & Manual Prompt Tuning \\
GPT4Graph\cite{GPT4Graph}   & Graph-to-text &+   & GPT-3                                                                               & LM           & Manual Prompt Tuning + Automatic Prompt Tuning    \\
Graph-LLM\cite{chen2023exploring}   & Graph-to-text &+   & \begin{tabular}[c]{@{}l@{}}BERT, DeBERTa, Sentence-BERT,\\ GPTs, LLaMA\end{tabular} & MLM,LM       & Manual Prompt Tuning + Automatic Prompt Tuning    \\ 
\bottomrule
\end{tabular}}

\label{tab:llm}
\end{table*}
% \usepackage{booktabs}
\begin{table*}[ht]
\centering
\caption{\footnotesize{Details of approaches involved as GNN+LLM-based models.} }
\resizebox{0.95\linewidth}{!}{
\begin{tabular}{@{}llll@{}}
\toprule
Model                        & Backbone Architecture             & Pre-training                  & Adaptation                                 \\ \midrule
% SimTeG \cite{SimTeG} & GNN-centric & MLM, TTCL & Parameter-Efficient FT \\ 
TAPE \cite{TAPE} & GNN-centric & LM & Tuning-free Prompting + Parameter-Efficient FT      \\ 
GIANT  \cite{GIANT}                      & GNN-centric                      & MLM                          & Vanilla FT                                   \\ 
GraD \cite{GraD}                        & GNN-centric                      & MLM                          & Parameter-Efficient FT                                    \\
GALM \cite{graphaware} & GNN-centric & Graph Reconstruction & Vanilla FT \\
OFA \citep{ofa} & GNN-centric & MLM & Tuning-free Prompting \\
Heterformer \citep{heterformer} & GNN-centric & LM & Vanilla FT \\
edgeformers \citep{edgeformers} & GNN-centric & LM & Vanilla FT \\
LLMRec \citep{llmrec} & GNN-centric & LM & Tuning-free Prompting + Parameter-Efficient FT      \\ 
WalkLM \citep{walklm} & GNN-centric & MLM & Vanilla FT\\
% TouchUp-G \citep{touchup-g} & GNN-centric & MLM+LM & Vanilla FT \\
METERN \citep{metern} & GNN-centric & MLM & Parameter-Efficient FT \\
% ZeroG \citep{zerog} & GNN-centric & MLM & Prompt Tuning\\
LLM-GNN \citep{llm-gnn} & GNN-centric & LM & Parameter-Efficient FT\\
WTGIA \citep{WTGIA} & GNN-centric & LM & Parameter-Efficient FT\\
GHGRL \citep{GHGRL} & GNN-centric & LM & Vanilla FT \\
% OpenGraph \citep{opengraph} & GNN-centric & LM & Prompt Tuning \\
GLEM  \cite{GLEM}                       & Symmetric                         & MLM                          & Vanilla FT                                   \\
% ConGrat     \cite{congrat}                 & Symmetric                         & MLM + GTCL               & Parameter-Efficient FT                                    \\
GraphFormer  \cite{GraphFormer}                & Symmetric                         & MLM                          & Vanilla FT                                  \\
G2P2 \cite{G2P2} &Symmetric&GTCL&Prompt Tuning\\
% SAFER    \cite{SAFER}                    & Symmetric                         & MLM                          & Parameter-Efficient FT                                    \\
Text2Mol   \cite{text2mol}                  & Symmetric                         & MLM + GTCL                             & Parameter-Efficient FT                                 \\
% MoMu      \cite{MoMu}                   & Symmetric                         & MLM + GTCL                   & Parameter-Efficient FT                          \\
MoleculeSTM  \cite{MoleculeSTM}                & Symmetric                         & MLM  + GTCL   & Parameter-Efficient  FT                                \\
MolCA \citep{molca} & Symmetric & LM & Parameter-Efficient FT \\
CLAMP   \cite{CLAMP}                     & Symmetric                         & MLM + GTCL               & Parameter-Efficient FT                             \\
GIT-Mol\citep{git-mol} & Symmetric & LM & Prompt Tuning \\
PATTON \citep{patton} & Symmetric & MLM & Parameter-Efficient FT\\
% InstructMol \citep{instructmol} & Symmetric& LM & Instruction Tuning\\
% GraphAdapter \citep{adaptergnn} & Symmetric & MLM  & Prompt Tuning \\
ENGINE \citep{engine} & Symmetric & LM & Parameter-Efficient FT\\
OpenGraph \citep{opengraph} & Symmetric & LM & Vanilla FT\\
RLMRec \citep{rlmrec} & Symmetric & LM & Parameter-Efficient FT\\
% Grenade \citep{grenade} & Symmetric & MLM+CL & Parameter-Efficient FT\\
% Graph-Toolformer \cite{zhang2023graph}            & LLM-centric                       & LM                           & Tuning-free Prompting + Vanilla FT\\ 
GraphTranslator \citep{graphtranslator} & LLM-centric & LM & Parameter-Efficient FT \\
THLM \citep{thlm} & LLM-centric  & MLM & Parameter-Efficient FT \\
% G-Prompt \citep{g-prompt} & LLM-centric & MLM & Prompt Tuning \\
GraphGPT \citep{graphgpt} & LLM-centric & MLM & {Prompt Tuning}\\
% GNNAVI & LLM-centric & LM & Parameter-Efficient FT\\
 InstructGraph\citep{instructgraph} & LLM-centric & LM & Prompt Tuning \\
 RELM \citep{relm} & LLM-centric & LM & {Tuning-Free Prompting}  \\
 GraphPrompter \citep{Graph-Prompter} & LLM-centric & LM & Parameter-Efficient FT \\
HiGPT \citep{tang2024higpt} & LLM-centric & LM & Parameter-Efficient FT \\
G-Retriever\citep{G-Retriever} & LLM-centric & LM & Prompt-Tuning \\
TEA-GLM\citep{TEA-GLM} & LLM-centric & LM & Parameter-Efficient FT \\
HyperBERT\citep{hyperbert2024} & LLM-centric & LM & Parameter-Efficient FT\\
\bottomrule                     
\end{tabular}}
\label{tab:llm+GNN}
\end{table*}

The success of foundation models depends on high-quality training data, and foundation models exhibit significantly different performance on various types of test data. In this section, we discuss the impact of graph data on graph foundation models from three aspects: graph type, graph scale and graph diversity.

\textbf{Graph Type}. Based on the number of categories of nodes and edges in a graph, we can categorize graphs into homogeneous graphs and heterogeneous graphs. In homogeneous graphs, all nodes and edges belong to the same category. For example, in a social graph where nodes represent individuals (users) and edges represent friendship relationships, it is a homogeneous graph because all nodes are individuals and all edges represent friendship relationships. Heterogeneous graphs, on the other hand, have more than one type of nodes or edges, representing different types of entities and relationships~\cite{shi2016survey}. For instance, an e-commerce graph may include nodes for users, products, and purchase relationships, forming a heterogeneous graph. For GFMs, handling heterogeneous graphs poses greater challenges and typically requires the design of specific backbone architectures and optimization objectives. Nonetheless, utilizing the meta-path based approach~\cite{wang2019heterogeneous}, a heterogeneous graph can be mapped into multiple homogeneous graphs, one for each meta-path. For example, one can apply the GFM trained on homogeneous graphs to each of these meta-path induced homogeneous graphs, separately, to get node embedding. Then, these embeddings on homogeneous graphs under different meta-paths can be fused together. However, beyond homogeneous graphs and heterogeneous graphs, there are some more complex types of graphs in the real world, such as dynamic graphs and hypergraphs~\cite{thomas2022graph}, which poses additional challenges for GFM.

\textbf{Graph Scale}. Based on the number of nodes and edges in a graph, we can categorize graphs into relatively small graphs and large graphs. Small graphs are of smaller scale, typically containing dozens to hundreds of nodes and edges. For example, chemical molecular graphs represent the structure of small molecules and typically consist of dozens to hundreds of atoms. Large graphs, on the other hand, refer to graphs with a significant number of nodes and edges, often encompassing millions or even billions of nodes and edges. For instance, e-commerce graph in Alibaba includes billons of nodes and hundreds of billion edges~\cite{zhang2020agl}. 
For graph foundation models, large graphs impose higher demands on the capacities of graph foundation models. Firstly, large graphs, due to their numerous nodes and typically sparser edges, introduce more noise and pose greater challenges in terms of storage and computation~\cite{yang2019aligraph}. Additionally, large graphs often exhibit long-range dependency relationships~\cite{dwivedi2022long}, requiring more neural network layers and a higher number of parameters, which exacerbates the over-smoothing~\cite{li2018deeper} and over-squashing~\cite{alon2020bottleneck} problem of GNN-based models.

\textbf{Graph Diversity}. Based on whether a graph dataset originates from the same domain, we can categorize graphs into same-domain graphs and cross-domain graphs. Same-domain graphs refer to graph data from similar or related domains, typically containing nodes and edges of similar types. For example, the social graphs of Facebook and WeChat come from similar domains. Cross-domain graphs~\cite{zhu2021cross}, on the other hand, involve graph data from different domains or data sources, often comprising nodes and edges of different types, aimed at addressing multi-domain problems or cross-domain tasks. For example, academic networks and molecular graphs come from different domains. 
For graph foundation models, supporting cross-domain graphs presents greater challenges because graphs from different domains lack a unified underlying semantics. This can result in weak transfer performance or even negative transfer when applying the model to a new dataset~\cite{zhou2022mentorgnn}. Therefore, addressing the heterogeneity of different domains and enabling the same GFM to be applied to graphs from different domains is a significant challenge for GFMs.

\subsection*{B. Impact of Graph Tasks on GFMs}
\label{sec:impact-graph-task}
Language foundation models can be widely applied to various NLP tasks, while for graph foundation models, the formats of graph tasks are also quite diverse and can be categorized into three classes: node-level tasks, edge-level tasks, and graph-level tasks.

\textbf{Node-level Tasks}. Node-level tasks refer to the classification, regression, or prediction performed on each node. Common node-level tasks include node classification, node regression, and clustering coefficient prediction. For example, in social networks, graph nodes can represent users, and node classification can be used to identify users from different social circles.

\textbf{Edge-level Tasks}. Edge-level tasks involve the classification, regression, or prediction performed on each individual edge. Common edge-level tasks include edge classification, link prediction, shortest path prediction, connectivity prediction, and maximum flow prediction. For example, in e-commerce, link prediction can be used to predict products that users may be interested in.

\textbf{Graph-level Tasks}. Graph-level tasks focus on the entire graph. Common graph-level tasks include graph classification, graph regression, graph generation, graph clustering, graph condensation and average clustering coefficient prediction. For example, in bioinformatics, graph property prediction can be used to predict the biological activity or toxicity of molecular compounds, thereby accelerating the drug discovery process.

In summary, the format of tasks in graphs are highly diverse and can be categorized into three types: node-level, edge-level, and graph-level, each of which has wide-ranging applications. This undoubtedly increases the challenge of homogenization for GFMs. 
For example, in graph classification and node classification tasks on synthetic datasets, modeling structural information is often more crucial~\cite{you2020design}. On the other hand, when dealing with node classification tasks on graphs with rich node features, modeling feature information becomes more important~\cite{you2020design}. 
Furthermore, tasks that are more similar to each other will also have a lower transfer difficulty, implying that these tasks are more likely to be addressed using the same GFM. While increasing expressive power holds promise for improving the performance of many node-level, edge-level, and graph-level tasks~\cite{li2020distance}, there is also some work suggesting that overly strong expressive power may not be necessary for graph generation tasks~\cite{zou2023will}.

\subsection*{C. Details of approaches involved as GNN-based models}\label{sec:gnn-table}
We categorize the GNN-based methods in Table~\ref{tab:gm}.
\subsection*{D. Details of approaches involved as LLM-based models}\label{sec:llm-table}
We categorize the LLM-based methods in Table~\ref{tab:llm}.
\subsection*{E. Details of approaches involved as GNN+LLM-based models}\label{sec:gnn+llm-table}
We categorize the GNN+LLM-based methods in Table~\ref{tab:llm+GNN}.

\end{document}